\documentclass{article}

 \usepackage[main, final]{neurips_2026}

\usepackage[utf8]{inputenc} 
\usepackage[T1]{fontenc}    
\usepackage{hyperref}       
\usepackage{url}            
\usepackage{booktabs}       
\usepackage{amsfonts}       
\usepackage{nicefrac}       
\usepackage{microtype}      
\usepackage[table]{xcolor}         

\usepackage{amsmath}
\usepackage{amssymb}
\usepackage{cleveref}
\usepackage{wrapfig}
\usepackage{graphicx}
\usepackage{caption}
\usepackage{ulem}
\usepackage{cancel}
\usepackage{subfig}
\usepackage{algorithm,algpseudocode}
\usepackage{placeins}
\usepackage{wasysym}

\crefname{equation}{Eq.}{Eqs.}
\crefname{section}{Sec.}{Secs.}
\crefname{table}{Tab.}{Tabs.}
\crefname{figure}{Fig.}{Figs.}
\crefname{algorithm}{Alg.}{Algs.}
\crefname{appendix}{App.}{Apps.}

\title{Distribution Matching Distillation\\without Fake Score Network}

\author{%
  Youngjoong Kim\textsuperscript{1}\quad
  Deokyeong Lee\textsuperscript{2}\quad
  Jaesik Park\textsuperscript{1}\thanks{Corresponding author: \texttt{jaesik.park@snu.ac.kr}}\\
  \textsuperscript{1}Department of Computer Science and Engineering, Seoul National University \\
  \textsuperscript{2}Department of Computer Science and Engineering, Sogang University \\
  \texttt{noah.kim@snu.ac.kr, plmft@sogang.ac.kr, jaesik.park@snu.ac.kr}
}

\begin{document}

\maketitle

\begin{abstract}
Distribution Matching Distillation (DMD) provides an effective distribution-level correction for few-step generation, while relying on an auxiliary fake-score network to track the evolving generative distribution. Recent work combines DMD with a flow-map objective to exploit the benefits of both methods: mode-seeking and mode-covering behaviors. Since this fake-score network increases memory and update overhead, we study whether this explicit tracker can be avoided when the generator has a flow-map structure. In this work, we propose \textit{Fake-Score-network-Free DMD (FSF-DMD)}, a DMD-style correction for flow-map generators, replacing the auxiliary fake-score estimator with a generator-induced surrogate. We observe that this surrogate can be derived from the definition of the fake velocity under a certain perturbation assumption. Building on this insight, we derive a DMD-style objective without a fake-side network, extend it to backward simulation for flow maps, and to a self-teacher variant for training from scratch. In our ImageNet-1K $256\!\times\!256$ experiments, FSF-DMD is competitive with the studied DMD baselines when initialized from pretrained networks and when trained from scratch.
\end{abstract}

\section{Introduction}
\label{sec:introduction}

Diffusion models~\citep{NEURIPS2020_4c5bcfec, song2021scorebased} and flow-matching models~\citep{lipman2023flow, liu2023flow} have achieved remarkable progress in high-fidelity generation, while relying on multistep sampling. To reduce generation cost, prior work has focused on few-step generation through flow-map models~\citep{boffi2025flow, geng2026mean, kim2024consistency, pmlr-v202-song23a} and Distribution Matching Distillation (DMD)~\citep{yin2024improved, yin2024onestep}.

These methods are based on two complementary paradigms: forward and reverse divergences. Flow-map generators are often trained with forward divergence, which uses data samples to compute loss objectives. These models learn the transport map along the probability flow and are known to exhibit mode-covering behavior~\citep{zheng2026large}. In contrast, DMD and its variants are based on reverse divergence, using generated samples to compute objectives. This can provide a mode-seeking correction for the generative distribution. To take advantage of both paradigms, recent studies jointly train the network with both objectives~\citep{cheng2026twinflow, ge2026senseflow, zheng2026large}.

DMD methods provide distribution-level correction by comparing the scores of the data distribution (teacher score) and the generative distribution (fake score). However, they rely on an auxiliary network to track the fake score during training. This tracker increases memory overhead and often requires additional update heuristics to stabilize training~\citep{ge2026senseflow, yin2024improved}. This raises a question: \textit{is the fake-score network necessary for DMD-style objectives in flow-map generators?}

In this work, we study this question in the setting where flow-map-style and DMD-style objectives are jointly trained. By analyzing the properties of flow maps, we observe that \textit{the endpoint pseudo-velocity of a flow-map generator can provide a tractable surrogate for fake-velocity estimation}. This motivates us to replace the auxiliary fake-side estimator in DMD with a generator-induced surrogate.

We introduce \textit{Fake-Score-network-Free DMD (FSF-DMD)}, which keeps the DMD perspective of a fake-side correction, while replacing the explicit fake-score network with our surrogate. We also extend it to backward simulation for flow maps, and to a self-teacher variant for training from scratch.

This replacement has several practical implications. (i) Since the fake-side surrogate is computed from the current generator, it changes with the evolving generator without requiring a separate training loop. (ii) By exploiting the semigroup property of flow-map generators, it can yield reasonable performance even without backward simulation for distribution matching. (iii) Finally, the generator is not trained to separately predict the fake velocity, and its capacity remains focused on generation.

We evaluate FSF-DMD on ImageNet-1K $256\!\times\!256$ and CelebA-HQ in few-step generation settings. Our experiments are intended as a proof of concept for the proposed network-free replacement, providing evidence for its practical effectiveness across the studied initialization conditions. In the flow-map-initialized comparison, FSF-DMD reaches an FID lower than or equal to those of the listed consistency distillation and DMD2 comparisons, without an explicit fake-score network. We further evaluate FSF-DMD under flow-matching initialization and even in a training-from-scratch setting, where the tracker-free correction remains effective in the studied settings.

\begin{figure}[t]
  \centering
  \includegraphics[width=\columnwidth]{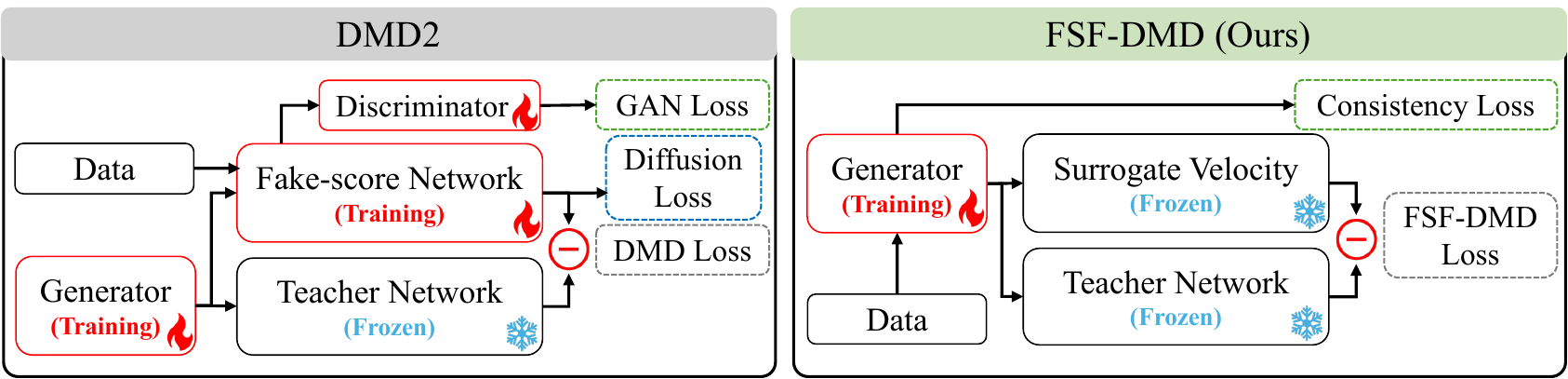}
  \caption{\textbf{Comparison between DMD2 and FSF-DMD.} (Left) DMD2 computes the discrepancy between the teacher and fake scores of the generated samples. It requires an additional network to estimate the fake score, introducing overhead through a separate training loop. (Right) For flow-map generators, FSF-DMD replaces the fake-side estimator with the generator-induced surrogate in \cref{eq:surrogate}, avoiding the additional fake-velocity network and secondary training loop.}
  \label{fig:overview}
\end{figure}

\section{Related Work}
\label{sec:related-work}

\textbf{Diffusion and Flow Models.} Diffusion~\citep{NEURIPS2020_4c5bcfec, song2021scorebased} and flow models~\citep{lipman2023flow, liu2023flow} are generative models that iteratively transform a prior distribution into a data distribution. They have achieved remarkable success in high-fidelity generation, while their multistep sampling limits the generation speed.

\textbf{Flow-Map Models.} To improve the sampling efficiency of these models, Consistency Models~\citep{pmlr-v202-song23a} directly map any point along a flow trajectory to its endpoint. Consistency Trajectory Models~\citep{kim2024consistency} and Flow Map Matching~\citep{boffi2025flow} generalize this concept by enabling mappings between any two arbitrary timesteps, called flow maps. MeanFlow~\citep{geng2026mean, geng2025improvedmeanflowschallenges} leverages the average velocity to predict the integral of the ODE, which coincides with the flow map. Most flow-map models are governed by forward-divergence objectives, which are defined on real data and are known to encourage mode-covering. These objectives often spread out densities and may generate samples from low-likelihood regions.

\textbf{Distribution Matching Distillation (DMD).} As a reverse-divergence method, DMD~\citep{yin2024improved, yin2024onestep} matches generated samples to the data distribution by comparing their denoising scores on noised distributions. It typically uses auxiliary networks to estimate the score of the generated distribution, which increases memory pressure. Additionally, it exhibits mode-seeking behavior and is known to reduce the diversity of generated samples compared to forward divergence~\citep{Lu_2025_ICCV}.

Recently, several approaches~\citep{cheng2026twinflow, ge2026senseflow, zheng2026large} jointly incorporate flow-map-style and DMD-style objectives to cover both mode-covering and mode-seeking behaviors. They have been successfully deployed on large-scale models, and we summarize their design choices in \cref{tab:related-design}.

To alleviate the memory overhead of a separate fake-score network, recent methods have explored self-contained approximations. TwinFlow~\citep{cheng2026twinflow} jointly trains the generator network to estimate the fake score. This refines generated samples without an explicit fake-side network, while an expressivity overhead in the generator network may arise. Moment Matching Distillation~\citep{salimans2024multistep} approximates the distributional difference via a first-order expansion in parameter space, thereby minimizing the distributional gap without an explicit fake-side network. However, relying on a first-order approximation may introduce truncation errors.

\begin{table}[t]
    \centering
    \caption{\textbf{Design choices of DMD-style objectives combined with flow-map generators.} \textit{Base flow map} denotes the flow-map objective jointly trained with DMD, \textit{GAN} denotes whether a GAN loss is adopted, \textit{Fake-side estimator} denotes the fake-score or fake-velocity estimator, and \textit{Rollout} denotes the few-step generation strategy used during training. FSF-DMD uses the generator-induced surrogate velocity and extends backward simulation to flow-map models based on the Euler method.}
    \label{tab:related-design}
    \vspace{1em}
    \resizebox{0.95\textwidth}{!}{%
    \begin{tabular}{llcll}
        \toprule
        Method & Base flow map & GAN & Fake-side estimator & Rollout \\
        \midrule
        DMD2~\citep{yin2024improved} & - & $\ocircle$ & Auxiliary network & Backward simulation \\
        rCM~\citep{zheng2026large} & sCM~\citep{lu2025simplifying} & $\times$ & Auxiliary network & rCM rollout \\
        SenseFlow~\citep{ge2026senseflow} & ISG~\citep{ge2026senseflow} & $\ocircle$ & Auxiliary network & Backward simulation \\
        TwinFlow~\citep{cheng2026twinflow} & RCGM~\citep{sun2026anystep} & $\times$ & Twin velocity & One-step generation \\
        \midrule
        \textbf{FSF-DMD (Ours)} & iMF~\citep{geng2025improvedmeanflowschallenges} & $\times$ & Surrogate velocity (Eq.~\ref{eq:surrogate}) & Euler rollout (Eq.~\ref{eq:tilde-F}) \\
        \bottomrule
    \end{tabular}
    }
\end{table}

\section{Method}
\label{sec:method}

In this section, we propose \textit{Fake-Score-network-Free DMD (FSF-DMD)} for flow-map generators. We review flow matching, flow maps, and distribution matching distillation. We then use the flow-map structure to replace the fake-score network with a generator-induced surrogate, as presented in \cref{fig:overview}.

\subsection{Preliminary}

\textbf{Notation.} We use $s$, $v$, $X$, and $u$ for scores, velocities, flow maps, and pseudo-velocities, respectively. Subscripts $\theta$, $\Phi$, and $\psi$ denote the parameters of the main model, teacher, and fake-side networks. Unparameterized quantities such as $v_\mathrm{fake}$ denote ideal targets. A hat denotes a generated sample, and the subscript $t$ denotes its noise level. We summarize the notation in \cref{tab:notation}.

\textbf{Flow Matching} models learn a velocity field from an interpolant that transports particles between a prior distribution and the target data distribution. Let $p_\mathrm{real}$ denote a data distribution. For a tractable prior distribution $p_Z$, previous work~\citep{lipman2023flow, liu2023flow} defines the linear interpolation $x_t = (1-t)x + tz$, where $x\sim p_\mathrm{real}$, $z\sim p_Z$, and $t\in [0, 1]$.

The conditional velocity along this path is $v_t(x_t|x) = (x_t - x) / t$, and the marginal velocity is given by $v^*_t(x_t) = \mathbb E_{x|x_t}[v_t(x_t|x)]$. The conditional flow-matching objective, $\mathcal L_\mathrm{CFM}$, guides the network $v_\theta$ to learn this velocity field, $v_\theta(x_t; t)\approx v^*_t(x_t)$:
\begin{align}
    \mathcal L_\mathrm{CFM} = \mathbb E_{x\sim p_\mathrm{real},z\sim p_Z, t\sim \mathcal U[0, 1]}[\|v_\theta(x_t; t) - v_t(x_t|x)\|^2_2].
\end{align}
Generation follows the probability-flow ODE (PF-ODE) induced by $v^*_t$, namely $dx_t = v^*_t(x_t)\,dt$. Starting from $z\sim p_Z$ at $t=1$, the sample $\hat x$ at $t=0$ can be approximated with the Euler rollout $x_{t_{i-1}} = x_{t_i} + (t_{i-1} - t_i)v_\theta(x_{t_i}; t_i)$ for $i > 0$. Here, $\{t_i\}_{i=0}^N$ is a time discretization, $t_0 = 0$, $t_N = 1$, and $t_{i} < t_{i+1}$ for all $i<N$. A good approximation requires multiple sampling steps, which motivates a direct few-step generator.

\begin{table}[t]
\centering
\caption{\textbf{Notation.} Greek subscripts denote network parameters and other symbols denote ideal targets. A hat denotes a generated sample, and the subscript $t$ denotes its noise level.}
\label{tab:notation}
\vspace{1em}
\resizebox{\linewidth}{!}{%
\begin{tabular}{lll}
\toprule
Symbol & Description & Status \\
\midrule
$v^*_t$, $X$ & Marginal velocity and flow map of the probability-flow ODE. & Ideal \\
$v_\mathrm{fake}$, $v^\mathrm{ODE}_\mathrm{fake}$ & Ideal fake velocities under independent and PF-ODE perturbations. & Ideal \\
$s_\Phi$, $v_\Phi$ & Teacher-score and teacher-velocity networks (subscripted by $\Phi$). & Frozen \\
$s_\psi$, $v_\psi$ & Fake-score and fake-velocity networks (subscripted by $\psi$). & Training \\
$X_\theta$, $u_\theta$ & Flow-map generator and pseudo-velocity network: $G_\theta(z) = X_\theta(z; 1, 0)$. & Training \\
$\bar u_\theta$ & Pseudo-velocity accumulated during backward simulation. & Output \\
\bottomrule
\end{tabular}
}
\end{table}

\textbf{Flow Maps.} To construct a few-step generator from the underlying flow, a flow map~\citep{boffi2025flow, kim2024consistency} defines a mapping between two points $x_t$ and $x_s$ on the same trajectory:
\begin{align}\label{eq:flow-map}
    X(x_t; t, s) = x_t + \int^s_tv^*_\tau(x_\tau)d\tau.
\end{align}
Recent work~\citep{boffi2025flow} views this as a unifying object that consistency models~\citep{pmlr-v202-song23a, frans2025one} aim to learn. The flow map has useful properties, including injectivity and invertibility, when $v^*_t(x_t)$ is Lipschitz continuous. We use these properties to derive our fake-velocity surrogate in \cref{subsec:our-approach}. Following MeanFlow-style flow maps~\citep{geng2026mean, geng2025improvedmeanflowschallenges, sabour2026align}, we parameterize the flow-map network $X_\theta$ with the pseudo-velocity network $u_\theta$, given by $X_\theta(x_t; t, s) = x_t + (s-t)u_\theta(x_t; t, s)$, and the corresponding consistency objective is:
\begin{align}
    \mathcal L_\mathrm{CT}  &= \mathbb E_{x,z,t,s}\left[w_{t,s}\left\| u_\theta(x_t; t, s) - \operatorname{sg}\left[v_t(x_t|x) + (s-t)\frac{d}{dt}u_\theta(x_t; t, s)\right] \right\|^2_2\right]\label{eq:ct}
\end{align}
where $\operatorname{sg}[\cdot]$ is a stop-gradient operator and $w_{t,s}$ is a scalar weight. In particular, $u_\theta(x_t; t, s)$ represents the average velocity needed to move $x_t$ to time $s$, and $u_\theta(x_t;t,t)$ recovers the instantaneous velocity $v^*_t(x_t)$ in the $s\to t$ limit~\citep{geng2026mean}. We later use this parameterization to derive the fake-velocity surrogate without an additional network.

\begin{figure}[t]
\centering
\subfloat[Consistency distillation]{%
    \includegraphics[width=0.32\linewidth]{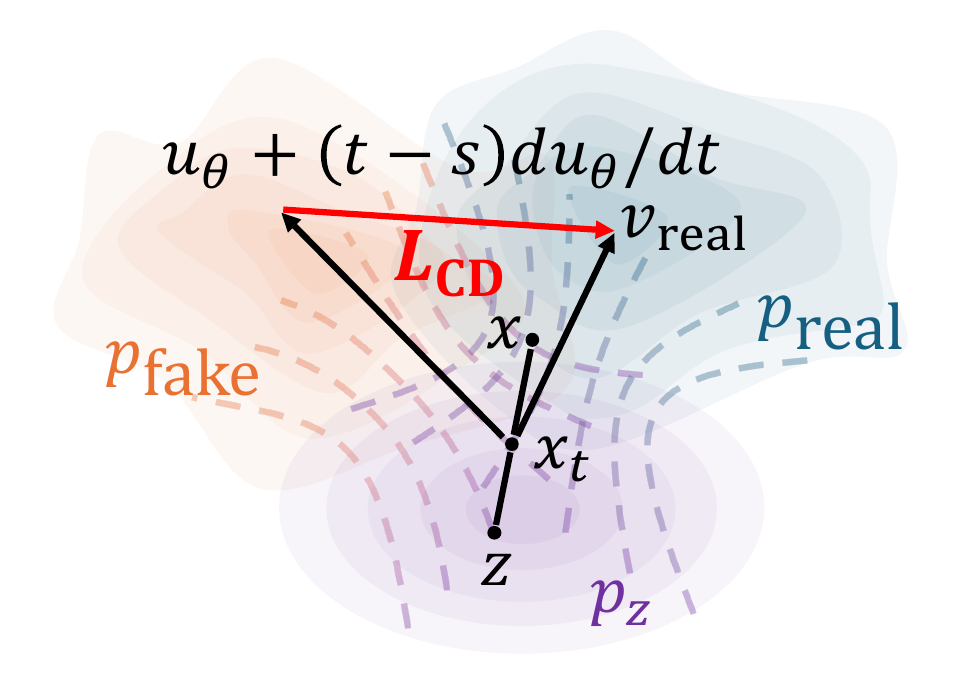}
    \label{subfig:cd}
}
\subfloat[DMD]{%
    \includegraphics[width=0.32\linewidth]{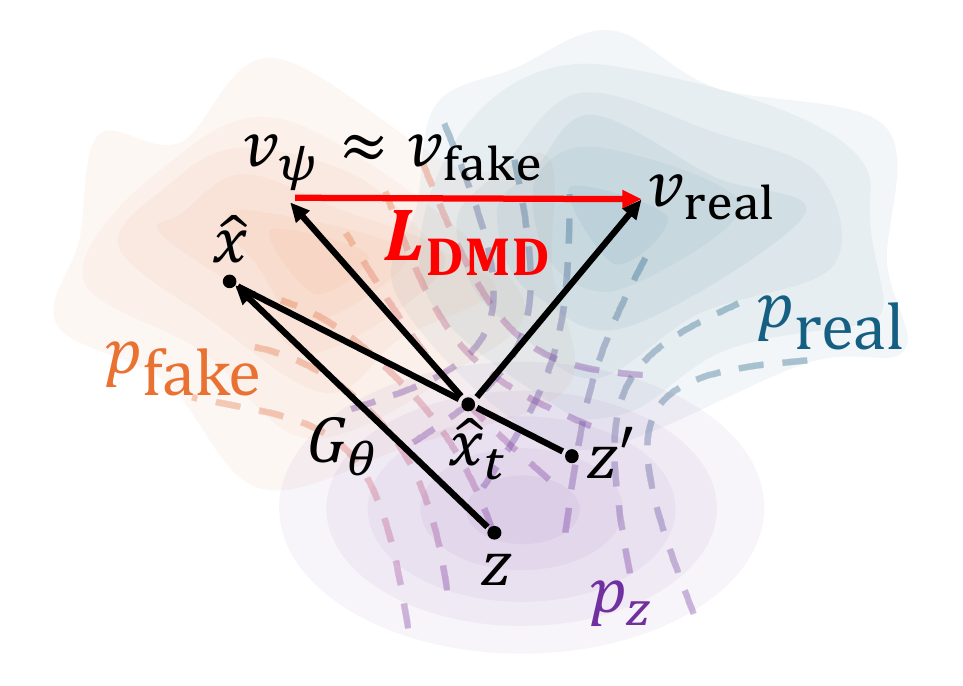}
    \label{subfig:dmd}
}
\subfloat[\textbf{FSF-DMD (Ours)}]{%
    \includegraphics[width=0.32\linewidth]{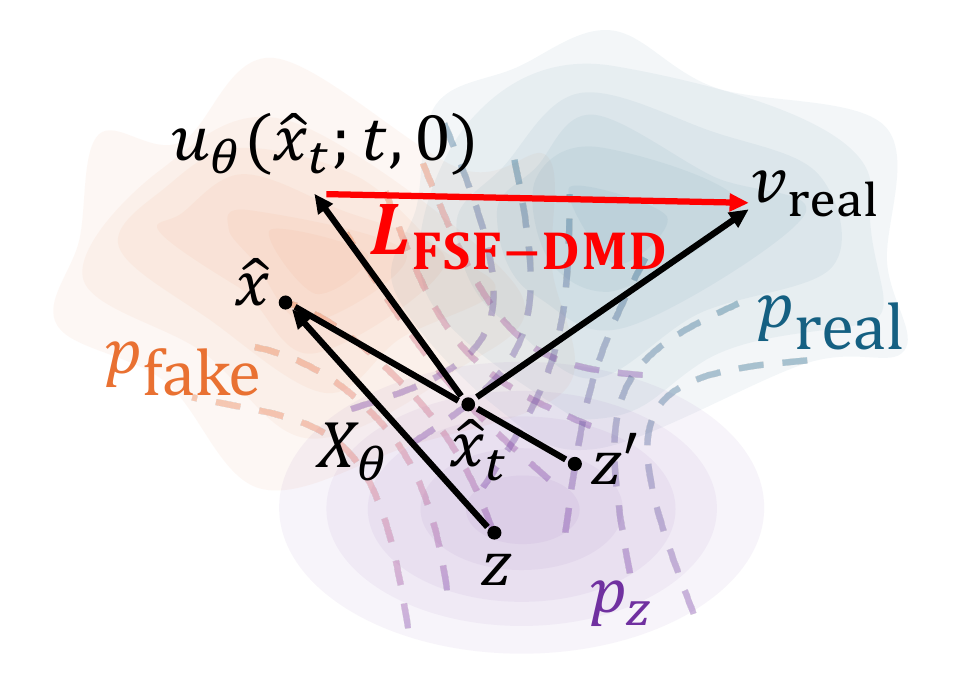}
    \label{subfig:fsfdmd}
}
\caption{\textbf{From explicit fake-score tracking to Fake-Score-network-Free DMD (FSF-DMD).} Red lines indicate the discrepancy measured by each objective. Consistency distillation $\mathcal L_\mathrm{CD}$ in \cref{eq:cd} anchors the generator to the flow-map structure, but does not include a fake-side correction (\cref{subfig:cd}). DMD applies a teacher-fake distributional correction $\mathcal L_\mathrm{DMD}$ in \cref{eq:dmd-score-correction}, but usually estimates the fake velocity with an auxiliary network $v_\psi$ (\cref{subfig:dmd}). Our $\mathcal L_\mathrm{FSF\text{-}DMD}$ in \cref{eq:fsf-dmd-impl} replaces $v_\psi$ with the pseudo-velocity $u_\theta(\hat x_t; t, 0)$ obtained from the flow-map generator $X_\theta$. We jointly optimize this objective with $\mathcal L_\mathrm{CD}$ to preserve the flow-map structure.}
\label{fig:teaser}
\end{figure}

\textbf{Distribution Matching Distillation (DMD)}~\citep{yin2024onestep} matches the real and generative distributions through the difference between their perturbed scores to train a few-step generator $G_\theta$. Let $p_{\mathrm{fake}} = (G_\theta)_\# p_Z$ be the push-forward generative distribution. DMD minimizes the Kullback-Leibler (KL) divergence between the noised distributions:
\begin{align}
    \nabla_\theta\mathcal L_\mathrm{DMD} = \mathbb E_{z,z',t}\left[(\nabla_x\log p_\mathrm{fake}(\hat x_t; t) - \nabla_x\log p_\mathrm{real}(\hat x_t; t))\frac{d\hat x_t}{d\theta}\right],
\end{align}
for $\hat x_t = (1-t)G_\theta(z) + tz'$ and $z'\sim p_Z$. In this case, we assume the teacher-score network $s_\Phi(x_t; t)\approx\nabla_x\log p_\mathrm{real}(x_t; t)$. As the fake score changes with $G_\theta$, standard DMD introduces an auxiliary network $s_\psi$ to estimate the score of the current generative distribution:
\begin{align}\label{eq:dmd-score-correction}
    \nabla_\theta\mathcal L_\mathrm{DMD} = \mathbb E\left[(s_\psi(\hat x_t; t) - s_\Phi(\hat x_t; t))\frac{d\hat x_t}{d\theta} \right] = \mathbb E\left[\gamma_t\left(v_\Phi(\hat x_t; t)-v_\psi(\hat x_t; t)\right)\frac{d\hat x_t}{d\theta}\right],
\end{align}
for $\gamma_t = (1-t)t^{-1}$. Under the score-velocity relation, the same gradient can be written in terms of $v_\Phi$ and $v_\psi$, the velocities corresponding to the two scores (\cref{subfig:dmd,app:score-velocity-connection}). The key point is that the fake score is not fixed: it changes as $G_\theta$ updates. Therefore, \textit{DMD uses a separate fake-score tracker, represented here by $s_\psi$ or $v_\psi$, which introduces additional computational overhead}.

These fake-side networks need to track the current generative distribution within a limited update budget, which can be challenging when the generator evolves rapidly. To stabilize training of fake-side networks, the Two Time-scale Update Rule (TTUR)~\citep{yin2024improved} and Implicit Distribution Alignment (IDA)~\citep{ge2026senseflow} are introduced. This suggests that estimating the fake score from the current generator itself can reduce the need for a separate tracker and stabilization designs, which is our goal.

\subsection{Our Approach}
\label{subsec:our-approach}

The accuracy of the fake-side tracker can affect the training dynamics, and maintaining this network introduces extra memory overhead during training. Our goal is not to improve the fake-velocity estimator itself, but to estimate the fake velocity without introducing an additional network.

\textbf{Motivating Observation.} This becomes tractable when the generator has the flow-map structure in \cref{eq:flow-map}. Starting from the flow-map setting, the auxiliary fake velocity $v_\psi$ would be trained by
\begin{align}\label{eq:fake-side-target}
    \mathcal L_\psi = \mathbb E_{z,z',t}\left[\left\|v_\psi(\hat x_t; t) - (z' - \hat x)\right\|^2_2\right],\quad v_\psi(\hat x_t; t)\approx \mathbb E_{z,z'|\hat x_t}[z' - z + u_\theta(z; 1, 0)],
\end{align}
where $\hat x = G_\theta(z) = X_\theta(z; 1, 0)$. If the perturbation reuses the generator input, $z'=z$, the fake velocity becomes \textit{a conditional expectation over the pseudo-velocity, $\mathbb E_{z|\hat x_t}[u_\theta(z; 1, 0)]$}. However, the conditional expectation is still not directly accessible since $p(z|\hat x_t)$ is nontrivial in general. This explains why a fake-score or fake-velocity network is needed to approximate it. This also shows that altering the perturbation process for $\mathcal L_\psi$ can provide a meaningful fake-side surrogate, motivating two steps: (i) express the fake velocity in terms of $u_\theta$, and (ii) simplify the conditional expectation.

\textbf{Deriving the Surrogate.} Switching from the prior-space posterior $p(z|\hat x_t)$ to the sample-space posterior $p(\hat x|\hat x_t)$, the fake-velocity target can be written as $v_\mathrm{fake}(\hat x_t; t) = \mathbb E_{\hat x|\hat x_t}[(\hat x_t - \hat x)/t]$. A further simplification appears under a PF-ODE-based noising process, which defines an inverse map $\hat x_t = X(\hat x; 0,t)$. By injectivity and invertibility of the flow map, the posterior collapses to a point mass $p(\hat x|\hat x_t) = \delta(X(\hat x; 0, t) - \hat x_t)$, yielding the corresponding fake velocity $v_\mathrm{fake}^\mathrm{ODE}$ (see \cref{app:deriving-fsf-dmd}):
\begin{align}
    v_\mathrm{fake}^\mathrm{ODE}(\hat x_t; t) &= \mathbb E_{\hat x|\hat x_t}[t^{-1}(\hat x_t - \hat x)] = t^{-1}(\hat x_t - X(\hat x_t; t,0)) && \because \text{PF-ODE coupling} \\
    &\approx t^{-1}(\hat x_t - X_\theta(\hat x_t; t, 0)) && \because \text{network approximation} \\
    &= u_\theta(\hat x_t; t, 0) && \because \text{definition of } u_\theta \label{eq:surrogate}
\end{align}
Therefore, under the PF-ODE coupling, the pseudo-velocity $u_\theta(\hat x_t; t,0)$ approximates the fake velocity $v_\mathrm{fake}^\mathrm{ODE}(\hat x_t; t)$ without introducing a separate network or training loop. Since it is computed from the generator, it reflects the evolving generative distribution.

\textit{We use it as a tractable surrogate for the fake velocity, replacing the explicit fake-velocity estimator}:
\begin{align}\label{eq:fsf-dmd-ideal}
    \Delta_{u, \Phi}(\hat x_t,t) = u_\theta(\hat x_t; t,0) - v_\Phi(\hat x_t; t),\quad \nabla_\theta\mathcal L(u, \Phi) = \mathbb E_{z,z',t}\left[-\gamma_t\Delta_{u, \Phi}(\hat x_t, t)\frac{d\hat x_t}{d\theta}\right].
\end{align}
In the landscape of self-contained approximations discussed in \cref{sec:related-work}, our formulation uses an endpoint evaluation of the same flow-map generator. It therefore avoids adding a separate fake-velocity objective or branch to the generator, keeping its capacity focused on the generation task.

\textbf{Field Inconsistency.} We note that the fake velocity in \cref{eq:fake-side-target} is determined by the noising process used to construct $\hat x_t$. DMD uses the independent perturbation for both teacher and fake velocities: $\hat x_t = (1-t)\hat x + tz'$ for $\hat x \perp z'$. The corresponding flow fields are therefore consistently defined. However, assuming PF-ODE noising while retaining $\hat x_t$ can induce inconsistency between the two fields, $v_\mathrm{fake}\ne v^\mathrm{ODE}_\mathrm{fake}$, as observed in \cref{tab:vel-discrepancy}. \cref{eq:surrogate} is therefore a heuristic surrogate rather than a pointwise estimate of the usual fake velocity, but remains effective in the studied settings (\cref{sec:experiments}).

As one possible intuitive explanation, this could steer the per-point generative direction $u_\theta(\hat x_t; t, 0)$ toward the teacher velocity $v_\Phi$. This complements flow-map training without requiring the surrogate to be the exact fake velocity. Additionally, this correction could be extended to the few-step generation setting while reducing training-inference mismatch (Eq.~\ref{eq:tilde-F}). We did not observe unusual optimization behavior attributable to this change in our settings (\cref{fig:loss-curves}). Studying how this difference affects final generation performance would be a promising direction for future work.

We also considered using PF-ODE noising directly for $\hat x_t$ and identified some concerns: (i) It would increase optimization complexity due to $d\hat x_t/d\hat x \approx \nabla X_\theta(\hat x; 0, t)$, compared to that of the independent perturbation, $d\hat x_t/d\hat x = 1-t$. (ii) Using PF-ODE noising in \cref{eq:fsf-dmd-ideal} would introduce another inconsistency between the evaluation field and the teacher velocity, and adapting the teacher network to the same field would collapse the correction signal. Thus, we retain independent perturbation for teacher evaluation as a practical engineering choice. See \cref{app:consider-pf-ode,app:ablation-study} for detailed discussions.

\subsection{Implementation}
\label{subsec:implementation}

The implemented FSF-DMD objective inserts this surrogate into a stop-gradient objective, given by:
\begin{align}\label{eq:fsf-dmd-impl}
    \mathcal L_\mathrm{FSF\text{-}DMD}(u, \Phi) = \mathbb E_{z,z',t}\left[w_t\left\|u_\theta(z; 1, 0) - \operatorname{sg}[u_\theta(z; 1, 0) - \Delta_{u, \Phi}(\hat x_t, t)]\right\|^2_2\right],
\end{align}
where $w_t$ is a scalar. Since the surrogate is derived under the flow-map generator, we use consistency distillation~\citep{sabour2026align}, $\mathcal L_\mathrm{CD}$, to align the generator with the flow-map structure. The full objective combines this term with \cref{eq:fsf-dmd-impl}, weighted by a scalar $\lambda$:
\begin{align}
    \mathcal L_\mathrm{base} &= \mathcal L_\mathrm{CD} + \lambda\mathcal L_\mathrm{FSF\text{-}DMD}(u, \Phi), \label{eq:fsf-dmd-full}\\
    \mathcal L_\mathrm{CD} &= \mathbb E_{x,z,t,s}\left[w_{t,s}\left\|u_\theta(x_t; t,s) - \operatorname{sg}\left[v_\Phi(x_t; t) + (s-t)\frac{d}{dt}u_\theta(x_t; t, s)\right]\right\|^2_2\right]. \label{eq:cd}
\end{align}
In practice, FSF-DMD begins to diverge after 1K steps without $\mathcal L_\mathrm{CD}$, as shown in \cref{tab:fsfdmd-wo-cd}. This suggests that a meaningful endpoint pseudo-velocity depends on the flow-map structure of the generator. A detailed derivation is provided in \cref{app:deriving-fsf-dmd}, and the training algorithm follows \cref{alg:fsf-dmd-training}.

\textbf{Few-Step Generation.} In the few-step generation setting, except for the first step, the generator input is derived from the previously generated sample. This induces a training-inference mismatch when we directly use perturbed data samples during training, and may affect generation performance.

To minimize this mismatch, DMD2 introduces \textit{backward simulation}. It generates samples with the renoising strategy, $\hat x^{(j)}$, and perturbs them once more, $\hat x^{(j)}_{t_j}$, to construct the actual inputs:
\begin{align}\label{eq:dmd2-bwd-simulation}
    \hat x = G_\theta(\operatorname{sg}[\hat x^{(j)}_{t_j}]; t_j),\quad \hat x^{(i-1)} = G_\theta(\hat x_{t_i}^{(i)}; t_i),\quad \hat x^{(i)}_{t_i} = (1-t_i)\hat x^{(i)} + t_iz_i,\quad z_i\sim p_Z
\end{align}
where $\hat x^{(N-i)}$ is the $i$-th step generated sample, $\hat x^{(N)}_{t_N} = z$, and $j\sim \mathcal U[1, N]$. For simplicity, let $t_i = i/N$. However, unlike DMD2, the flow-map generator has a different generation pipeline:
\begin{align}
    \hat x_{i/N} = \hat x_{(i+1)/N} - \frac1N u_\theta\left(\hat x_{(i+1)/N}; \frac{i+1}{N}, \frac{i}{N}\right) = z - \frac1N\sum^N_{k=i+1}u_\theta\left(\hat x_{k/N}; \frac{k}{N}, \frac{k-1}{N}\right),
\end{align}
for $\hat x = \hat x_0$, $\hat x_1 = z$, and $0\le i < N$. We found that backward simulation has room for improvement in our setting, since it does not fully reflect the flow-map generation process. We therefore make it consistent with the flow-map generator. To manage the generation steps for inference and backward simulation separately, we use $N$ for inference-time generation steps and $M$ for backward-simulation steps.
Following \cite{huang2026self}, applying stop-gradient to intermediate samples gives:
\begin{align}\label{eq:tilde-F}
    \frac{d\hat x_0}{d\theta} = -\frac{d}{d\theta}\bar u_\theta(z; 1, 0),\quad \bar u_\theta(z; 1, 0) = \frac1M\sum^M_{i=1}u_\theta\left(\operatorname{sg}[\hat x_{i/M}]; \frac{i}M, \frac{i-1}{M}\right).
\end{align}
Since accumulating network outputs can increase memory pressure, we additionally apply stop-gradient according to the batch index $b\in [1, B]$: $\operatorname{sg}[u_\theta(\hat x_{i/M}; i/M, (i-1)/M)]$ for all $i \ne \lceil bM/B\rceil$. This provides room for additional memory optimization, and we observe consistent performance while reducing memory usage when it is applied (see \cref{app:ablation-study}). We use this setup when $M=2$. We note that this is optional rather than a necessary component of the contribution.

Incorporating \cref{eq:tilde-F} into the single-step objective in \cref{eq:fsf-dmd-impl} gives our final few-step objective:
\begin{align}
    \mathcal L_\mathrm{distill} &= \mathcal L_\mathrm{CD} + \lambda\mathcal L_\mathrm{FSF\text{-}DMD}(\bar u, \Phi)\label{eq:fsf-dmd-distill} \\
    \mathcal L_\mathrm{FSF\text{-}DMD}(\bar u, \Phi) &= \mathbb E_{z,z',t}\left[w_t\|\bar u_\theta(z; 1, 0) - \operatorname{sg}[\bar u_\theta(z; 1, 0) - \Delta_{u, \Phi}(\hat x_t, t)]\|^2_2\right].\label{eq:bwd-simulation}
\end{align}
This objective preserves the flow-map generation process and coincides with the initial version in \cref{eq:fsf-dmd-impl} when $M=1$. We observe that it gives better performance in our setting, as shown in \cref{tab:distillation-ablation}.

\textbf{Training from Scratch.} We can even extend FSF-DMD to the \textit{training-from-scratch setting} without a teacher network. Since the MeanFlow-style parameterization gives $u_\theta(x_t; t, t)\approx v^*_t(x_t)$, we can substitute the teacher network $v_\Phi(x_t; t)$ with the network approximation itself:
\begin{align}\label{eq:fsf-dmd-scratch}
    \mathcal L_\mathrm{scratch} &= \mathcal L_\mathrm{CT} + \lambda\mathcal L_\mathrm{FSF\text{-}DMD}(\bar u), \\
    \mathcal L_\mathrm{FSF\text{-}DMD}(\bar u) &= \mathbb E_{z,z',t}\left[w_t\|\bar u_\theta(z; 1, 0) - \operatorname{sg}[\bar u_\theta(z; 1, 0) - \Delta_u(\hat x_t, t)]\|^2_2\right], \\
    \Delta_u(\hat x_t, t) &= u_\theta(\hat x_t; t, 0) - u_\theta(\hat x_t; t, t).
\end{align}
For this setting, we replace consistency distillation with consistency training in \cref{eq:ct}. This objective provides a DMD-style correction for generated samples even in the training-from-scratch stage. It requires neither a teacher network nor an additional fake-velocity objective in generator training. We observe that this reverse-divergence surrogate can improve the flow-map generator in \cref{subsec:from-scratch}.

We finally use \cref{eq:fsf-dmd-distill} for the distillation setting, and \cref{eq:fsf-dmd-scratch} for the training-from-scratch setting. We use timestep shifting $\frac{\kappa t}{(\kappa - 1)t + 1}$~\citep{yao2025vavae,geng2026mean} and classifier-free guidance (CFG)~\citep{ho2021classifierfree} for conditional generation. Details are provided in \cref{app:experimental-details}.

In summary, FSF-DMD keeps the DMD view that distillation is driven by a teacher-fake distributional correction. Its difference is that the fake-side estimator is represented by pseudo-velocities from a flow-map generator, rather than by an auxiliary network or a separate fake-velocity objective. This lets the generator remain focused on generation while efficiently providing a distribution-matching signal. The next section evaluates this formulation and studies how our approach behaves in practice.

\section{Experiments}
\label{sec:experiments}

We evaluate whether our approach works in practice. The main results use ImageNet-1K~\citep{deng2009imagenet} at $256\!\times\!256$ resolution with VA-VAE latents~\citep{yao2025vavae}, and tables report FID50K~\citep{NIPS2017_8a1d6947}. We additionally report Inception Score (IS)~\citep{NIPS2016_8a3363ab} and precision/recall~\citep{NEURIPS2019_0234c510} in \cref{tab:additional-metrics}. We use the LightningDiT-B/1 architecture~\citep{yao2025vavae} throughout the ImageNet experiments and evaluate FSF-DMD across three training settings. Since FSF-DMD depends on the flow-map structure, we first study a distillation setting initialized from a pretrained flow-map network~\citep{geng2025improvedmeanflowschallenges}. We then relax the initialization to a pretrained flow-matching network~\citep{yao2025vavae}, and finally consider training from scratch with random initialization.

We assume 2-step generation ($N=2$), timestep shifting $\kappa=10$, $w_{t,s}=\cos(\pi t/2)$ for $\mathcal L_\mathrm{CD}$, and a CFG scale of 6.0 by default. For fair comparisons, we conduct ablation studies of DMD2 and its variants in the ImageNet setting over certain hyperparameters, weight initializations, TTUR, GAN loss, and backward simulation. We use official settings if possible and leave further details in \cref{app:experimental-details}.

\subsection{Proof of Concept: Flow-Map-Initialized Comparison}
\label{subsec:main-comparison}

\definecolor{myemerald}{rgb}{0.753, 0.898, 0.804}
\definecolor{mylightgreen}{rgb}{0.894, 0.933, 0.745}
\definecolor{myyellow}{rgb}{0.996, 0.972, 0.780}
\newcommand{\fst}{\cellcolor{myemerald!100}}
\newcommand{\snd}{\cellcolor{mylightgreen!100}}
\newcommand{\thd}{\cellcolor{myyellow!100}}
\begin{table}
    \centering
    \caption{\textbf{Comparison on ImageNet-1K in the flow-map-initialized setting.} \textit{Peak step} denotes the training step that achieves the reported FID, \textit{sec/step} denotes the wall-clock time per training step, and \textit{VRAM} denotes the VRAM requirement. The teacher achieves FID 8.54 with 8 sampling steps.}
    \label{tab:main-comparison}
    \vspace{1em}
    \resizebox{0.90\linewidth}{!}{%
    \begin{tabular}{lcccc}
        \toprule
        Method & FID & Peak step & sec/step & VRAM (GB) \\
        \midrule
        Consistency distillation~\citep{sabour2026align} & 6.06 & 36K & \fst\textbf{0.23} & \fst\textbf{26} \\
        DMD2~\citep{yin2024improved} & 4.18 & 15K & 1.46 & 60 \\
        DMD2 + IDA-style update~\citep{ge2026senseflow} & 4.20 & \thd 12K & 1.46 & 60 \\
        DMD2 + $\mathcal L_\mathrm{CD}$~\citep{zheng2026large} & 4.67 & 54K & 1.63 & 76 \\
        \midrule
        DMD2 + $\mathcal L_\mathrm{CD}$ + Random masking $s\gets t$ & \fst\textbf{3.85} & 63K & 1.68 & 76 \\
        \midrule
        \textbf{FSF-DMD (Ours, $M=2$)} & \fst\textbf{3.85} & \snd 4K & \thd 0.38 & \snd 49 \\
        \textbf{FSF-DMD (Ours, $M=1$)} & \thd 3.96 & \fst\textbf{3K} & \snd 0.31 & \snd 49 \\
        \bottomrule
    \end{tabular}
    }
\end{table}

\begin{wrapfigure}{R}{0.45\linewidth}
  \centering
  \includegraphics[width=\linewidth]{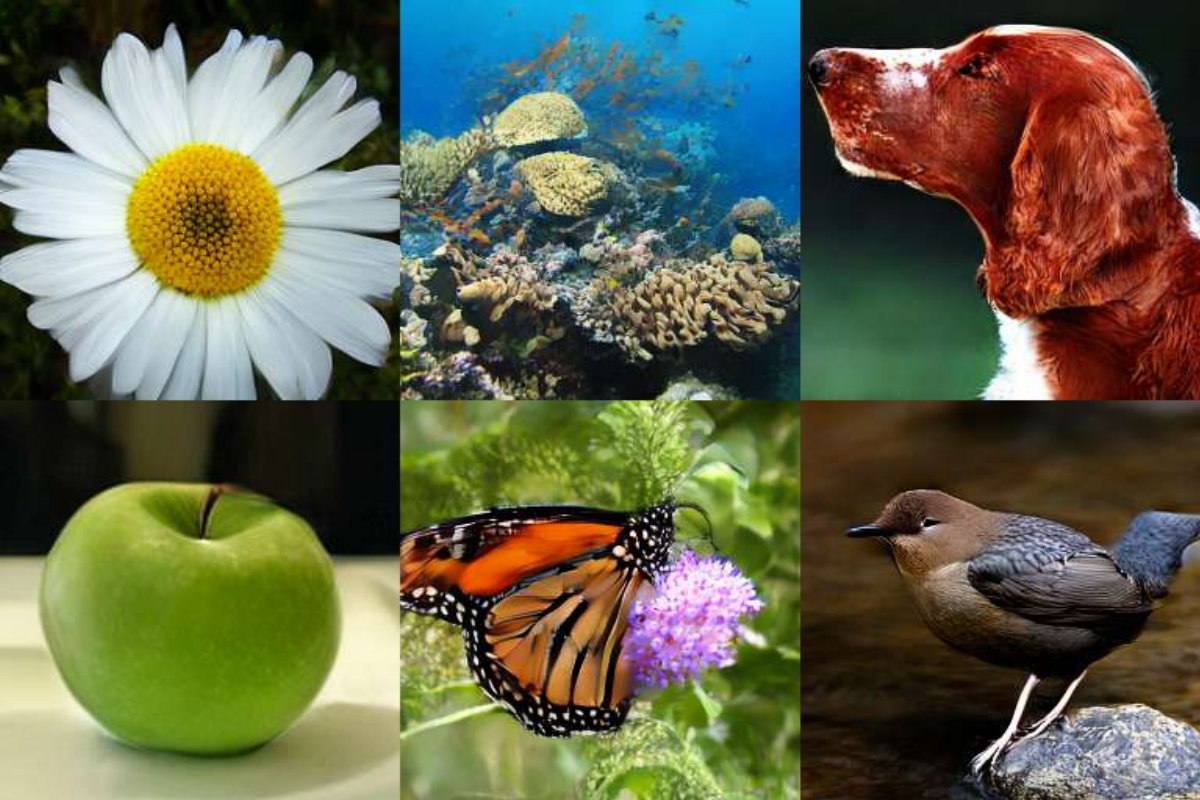}
  \caption{\textbf{2-step FSF-DMD samples} on ImageNet-1K $256\!\times\!256$, flow-map-initialized.}
  \label{fig:qual}
  \vspace{-3em}
\end{wrapfigure}

\cref{tab:main-comparison} compares FSF-DMD with consistency distillation and DiT-adapted DMD variants in the flow-map-initialized setting. For initialization, we use an improved MeanFlow model~\citep{geng2025improvedmeanflowschallenges} pretrained for 400K steps under our setup (FID 9.10). Reported methods peak before 70K steps, and we report the best FID found. Generated samples from FSF-DMD are shown in \cref{fig:qual}.

FSF-DMD improves the consistency distillation baseline from 6.06 to 3.85 FID. \textit{Under this controlled comparison, FSF-DMD obtains an FID equal to or lower than those of the listed variants without an explicit fake-score network or an adversarial branch.} It also reaches its best FID within 4K steps, earlier than the listed distillation variants in this setup.

These results provide proof-of-concept evidence that (i) the implementation-level surrogate is effective in practice, and (ii) FSF-DMD is competitive with explicit-network variants in this setting. Additional metrics in \cref{tab:additional-metrics} show that FSF-DMD also improves IS and recall in the flow-map-initialized setting.

This behavior may reflect two factors. First, our surrogate reflects the generator update, whereas a separate tracker must follow the generator through its own update schedule. Second, consistency distillation anchors the flow-map structure, which can provide a stable real-data signal without an adversarial branch.

Practically, FSF-DMD reduces the VRAM requirement and wall-clock time compared to the DMD2 variants on NVIDIA B200 GPUs, under the same local setup. \textit{The gap between DMD2 and FSF-DMD is mainly due to the additional fake-velocity network and TTUR, which FSF-DMD does not require.}

\textbf{Unconditional Generation.} We evaluate unconditional generation using CelebA-HQ at $256\!\times\!256$ scale with SD-VAE latents~\citep{Rombach_2022_CVPR} and the DiT-B/2 architecture, following \cite{frans2025one}. We adopt MeanFlow without CFG for flow-map initialization. We report the number of sampling steps $N$ and FIDs in \cref{tab:celeba-hq}. In this setting, FSF-DMD shows comparable results to the listed methods, remaining effective for unconditional generation. See \cref{fig:celeba-hq} for qualitative results.

\subsection{Ablation Study}
\label{subsec:distillation-ablations}

We study which components drive the result. As shown in \cref{tab:distillation-ablation}, the improvement appears when FSF-DMD is combined with backward simulation ($M=2$). Our flow-map version in \cref{eq:bwd-simulation} gives a better result than the DMD2-style simulation. Since it preserves the generation pipeline of the flow-map model, we attribute this gain to better alignment with the generation process.

\begin{table}[t]
\centering
\begin{minipage}{0.40\linewidth}
    \centering
    \caption{\textbf{Comparison on CelebA-HQ.}}
    \label{tab:celeba-hq}
    \vspace{1em}
    \resizebox{0.95\linewidth}{!}{%
    \setlength{\tabcolsep}{3.0pt}
    \begin{tabular}{lcr}
    \toprule
    Method & $N$ & FID \\
    \midrule
    Flow Matching~\citep{lipman2023flow} & 128 & 7.3 \\
    Shortcut Models~\citep{frans2025one} & 128 & 6.9 \\
        & 4 & 13.8 \\
    \midrule
    MeanFlow~\citep{geng2026mean} & 2 & 6.39 \\
        & 1 & 8.85 \\
    \midrule
    DMD2 ($M=2$) & 2 & 6.24 \\
    DMD2 ($M=1$) & 1 & 8.62 \\
    \textbf{FSF-DMD (Ours, $M=2$)} & 2 & \textbf{6.20} \\
    \textbf{FSF-DMD (Ours, $M=1$)} & 1 & 8.04 \\
    \bottomrule
    \end{tabular}
    }
\end{minipage}\hspace*{0.01\linewidth}
\begin{minipage}{0.515\linewidth}
    \centering
    \caption{\textbf{Ablation study on ImageNet-1K.}}
    \label{tab:distillation-ablation}
    \vspace{1em}
    \resizebox{\linewidth}{!}{%
    \begin{tabular}{lr}
        \toprule
        Distillation variant & FID \\
        \midrule
        Consistency distillation~\citep{sabour2026align} & 6.06 \\
        + FSF-DMD (Ours, Eq.~\ref{eq:fsf-dmd-impl}) & 5.70 \\
        + Backward simulation (DMD2, $M=2$) & 5.81 \\
        + Backward simulation (Ours, Eq.~\ref{eq:bwd-simulation}) & 5.34 \\
        + Hyperparameter tuning & 4.45 \\
        + Random masking $s\gets t$ & \textbf{3.85} \\
        \midrule
        FSF-DMD + Single-step simulation ($M=1$) & 3.96 \\
        DMD2 + Single-step simulation ($M=1$) & 4.78 \\
        \midrule
        Consistency distillation + Random masking & 4.98 \\
        \bottomrule
    \end{tabular}
    }
\end{minipage}
\end{table}

For hyperparameter tuning, we tune the scalar coefficients $\lambda$, $w_t$ for $\mathcal L_\mathrm{FSF\text{-}DMD}$, and the timestep-shifting parameter $\kappa$. The baseline setup, $\lambda=0.01$, $w_t = 1$, and $\kappa=1$ (no shifting), achieves an FID of 5.34. Adjusting them to $\lambda=0.1$, $w_t=1/|\bar u_\theta(z; 1, 0) - v_\Phi(\hat x_t; t)|$, and $\kappa=10$ improves the result to 4.45. Randomly masking $s$ to $t$~\citep{geng2026mean} further improves the final result to 3.85. We also evaluate consistency distillation with the same random masking. It reaches 4.98 FID, \textit{leaving a gap that supports the role of the DMD-style correction of FSF-DMD}. See \cref{app:ablation-study} for further studies.

Compared to the single-step simulation result for DMD2 in \cref{tab:distillation-ablation}, FSF-DMD with single-step simulation ($M=1$) shows a relatively small degradation. This is consistent with the semigroup property of flow maps, $X(x_t; t, s) = X(X(x_t; t, r); r, s)$, which allows backward simulation to be approximated by single-step generation. This behavior comes from the flow-map structure of the FSF-DMD generator, whereas DMD2 does not explicitly impose this structure.

FSF-DMD reaches an FID comparable to that of DMD2 + $\mathcal L_\mathrm{CD}$ with random masking in this setting. This is consistent with our intuition: the surrogate objective could steer the per-point generative direction toward the teacher direction, without requiring the surrogate to be the exact fake velocity.

\subsection{Relaxing to Flow-Matching Initialization}
\label{subsec:aux-ldit-teacher}

\begin{wraptable}{R}{0.36\textwidth}
    \centering
    \vspace{-1.2em}
    \caption{\textbf{Comparison on ImageNet-1K with flow-matching initialization.} The teacher is evaluated with 250-step Euler sampling.}
    \label{tab:ldit-teacher}
    \resizebox{0.36\textwidth}{!}{%
    \begin{tabular}{lr}
        \toprule
        Method & FID \\
        \midrule
        Teacher-velocity network & 12.44 \\
        + 3-channel CFG & 3.79 \\
        \midrule
        Consistency distillation & 5.53 \\
        DMD2 (TTUR=1:5)  & 5.53 \\
        DMD2 + $\mathcal L_\mathrm{CD}$ (TTUR=1:5) & 4.90 \\
        \midrule
        \textbf{FSF-DMD (Ours)} & \textbf{4.70} \\
        \bottomrule
    \end{tabular}
    }
    \vspace{-1em}
\end{wraptable}

We then relax the initialization from a pretrained flow-map model to a more common flow-matching model. We train LightningDiT-B/1~\citep{yao2025vavae} for 400K steps, and the FIDs reported in \cref{tab:ldit-teacher} are the best values we found under our setup. Unlike DMD2, consistency distillation and FSF-DMD require an additional $s$ condition, and we adopt the conditioning scheme of \cite{lee2026decoupled} without modifying the architecture for a fair comparison with the listed methods.

The results suggest that \textit{FSF-DMD can be trained under the relaxed initialization.} It improves over the consistency distillation baseline and the listed comparisons in this evaluation. In \cref{tab:additional-metrics}, FSF-DMD also achieves higher precision while keeping the recall on par with the listed methods in this setting.

Compared to \cref{tab:main-comparison}, initialization from a flow-map model gives better FSF-DMD results. This may be due to several factors. First, the generator is already capable of few-step generation, and the correction signal may be more reliable since the generated samples remain closer to the teacher distribution even at the early training stage. Second, the improved MeanFlow (iMF) model uses more parameters and computation, and part of the gap may simply reflect scaling. For these reasons, we next evaluate the training-from-scratch capability of FSF-DMD with the iMF baseline.

\subsection{From-Scratch Training}
\label{subsec:from-scratch}

\begin{figure}[th]
\begin{minipage}[th]{0.54\linewidth}
  \centering
  \includegraphics[width=\linewidth]{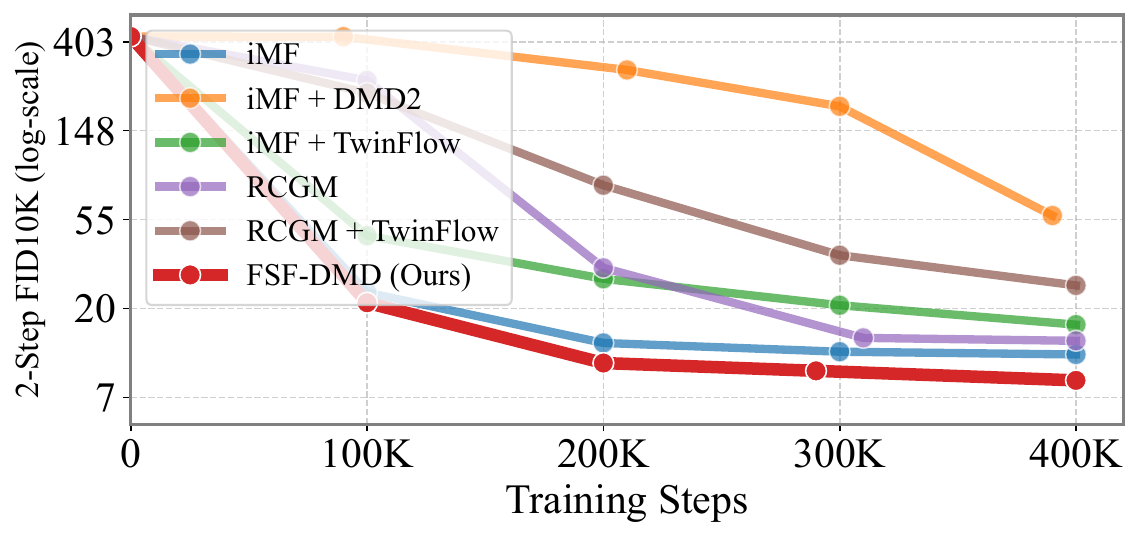}
  \caption{\textbf{FID10K over training steps} (log-scale).}
  \label{fig:loss-curves}
\end{minipage}\hspace*{0.01\linewidth}
\begin{minipage}[th]{0.445\linewidth}
    \centering
    \captionof{table}{\textbf{Training from scratch for 400K steps on ImageNet-1K} (FID50K and IS).}
    \label{tab:from-scratch}
    \resizebox{1.0\textwidth}{!}{%
    \begin{tabular}{lrr}
        \toprule
        Method & FID ($\downarrow$) & IS ($\uparrow$) \\
        \midrule
        iMF baseline~\citep{geng2025improvedmeanflowschallenges} & 9.10 & 148.71 \\
        \midrule
        iMF + DMD2 (TTUR=1:5) & 13.38 & 138.80 \\
        iMF + TwinFlow~\citep{cheng2026twinflow}\hspace*{-1.3em} & 12.17 & 144.26 \\
        \midrule
        RCGM~\citep{sun2026anystep} & 11.53 & 164.55 \\
        RCGM + TwinFlow & 22.58 & 106.04 \\
        \midrule
        \textbf{FSF-DMD (Ours)} & \textbf{6.35} & \textbf{194.91} \\
        \bottomrule
    \end{tabular}
    }
\end{minipage}
\end{figure}

We evaluate FSF-DMD from random initialization with \cref{eq:fsf-dmd-scratch}. In \cref{tab:from-scratch}, we augment the LightningDiT-B/1 network with in-context conditioning~\citep{geng2025improvedmeanflowschallenges} and train each network for 400K steps with single-step simulation ($M=1$). For TwinFlow~\citep{cheng2026twinflow}, we train a fake velocity with inverted timesteps, $\|u_\theta(\hat x_t; -t, -t) - (z' - \hat x)\|^2_2$, and apply DMD with this approximated fake velocity. Since TwinFlow is originally based on RCGM~\citep{sun2026anystep}, we report the RCGM results and use the official RCGM settings for both RCGM and TwinFlow.

As shown in \cref{tab:from-scratch}, FSF-DMD is not restricted to distilling pretrained models; it can be directly applied to train flow-map models entirely from scratch. \cref{fig:loss-curves} suggests that, compared to the other methods, training is faster and reaches a better result under the same training budget. One possible explanation is that FSF-DMD does not require the generator to learn a separate fake-velocity branch, allowing it to remain focused on generation.

In summary, the experiments support FSF-DMD as a corrective method without an explicit fake-score network for flow-map generators. The flow-map-initialized comparison suggests that the generator-induced surrogate can provide an effective correction in the studied setting. The relaxed initialization result suggests that the method is not tied to the strongest flow-map checkpoint. The from-scratch result suggests that the same correction can also help joint training without an external teacher.

We leave scaling to larger architectures and broader large-scale settings, including text-to-image generation, for future work.

\section{Conclusion}
\label{sec:conclusion}

We presented FSF-DMD, a fake-score-network-free DMD formulation for flow-map generators. The method keeps the DMD view that few-step generators benefit from a teacher-fake distributional correction, but replaces the auxiliary fake-score estimator with a generator-induced surrogate. This frames the flow-map structure not only as a forward-divergence training mechanism, but also as a way to provide the reverse-divergence signal without an explicit fake-score network. The resulting objective extends to flow-map-consistent backward simulation and to a self-teacher variant for training from scratch. The experiments support the effectiveness of this formulation in the studied settings. Overall, FSF-DMD suggests that flow-map generators can supply a useful fake-score correction in DMD while preserving the distribution-matching perspective.

\begin{ack}
This work was supported by the IITP grants (RS-2021-II211343: AI Graduate School Program (Seoul National University) (5\%), RS-2024-00509257: Global AI Frontier Lab (15\%), and RS-2026-25546560: Full-Cycle GenAI Talent Development - Safe, Accessible and Content-creating AI with an Interactive Education Platform (15\%)) and the NRF grants (RS-2024-00405857 (50\%), RS-2026-25618928 (15\%)) funded by the Korea government (MSIT).
\end{ack}

\bibliography{main}
\bibliographystyle{plain}


\newpage
\appendix

\section{Theoretical Analysis}
\label[appendix]{app:theoretical-analysis}

\subsection{Score-Velocity Connection}
\label[appendix]{app:score-velocity-connection}

Let $p_\mathrm{real}$ denote a data distribution. For a tractable prior distribution $p_Z$ and a time distribution $t\sim p_T$ satisfying $0\le t \le T$, $\alpha_t$ and $\sigma_t$ define the interpolation $x_t = \alpha_t x + \sigma_t z$, where $\alpha_0 = \sigma_T = 1$ and $\alpha_T = \sigma_0 = 0$. We assume that $\alpha_t$ is monotonically decreasing and $\sigma_t$ is monotonically increasing, with bounded first and second derivatives.

Throughout the paper, we assume $\alpha_t = 1 - t$, $\sigma_t = t$ for $T = 1$, and $p_Z = \mathcal N(0, I)$ by default.

For this stochastic interpolant, the conditional velocity is $v_t(x_t|x) = (x_t - x) / t = z - x$, and the marginal velocity is $v^*_t(x_t) = \mathbb E_{x|x_t}[v_t(x_t|x)]$. By definition, $v^*_t(x_t) = (x_t - \mathbb E_{x|x_t}[x])/t = (x_t - \mu_t^*(x_t)) / t$, where $\mu_t^*(x_t) = \mathbb E_{x|x_t}[x]$ is the posterior mean.

This stochastic interpolant gives $x_t|x\sim \mathcal N((1-t)x, t^2I)$. In this case, the conditional score is
\begin{align}
    s_t(x_t|x) &= \nabla\log p_t(x_t|x) = -\frac{x_t - (1-t)x}{t^2}, \\
    p_t(x_t|x) &= \frac{1}{(2\pi t^2)^{d/2}}\exp\left(-\frac{\|x_t - (1-t)x\|^2_2}{2t^2}\right).
\end{align}
The marginal score is defined by the conditional expectation of the conditional score function, $s^*_t(x_t) = \mathbb E_{x|x_t}[s_t(x_t|x)] = -(x_t - (1-t)\mu_t^*(x_t)) / t^2$.

Since both the marginal velocity and the marginal score share the term $\mu_t^*(x_t)$, we can derive their connection:
\begin{align}
    \mu_t^*(x_t) &= x_t - tv^*_t(x_t) \\
    s^*_t(x_t) &= -t^{-2} (x_t - (1-t)\mu^*_t(x_t))\\
    &= -t^{-2} (x_t - (1-t)(x_t - tv^*_t(x_t))) \\
    &= -t^{-2}(\cancel{x_t - x_t} + tx_t + tv^*_t(x_t) - t^2v^*_t(x_t)) \\
    &= -t^{-1}(x_t + (1-t)v^*_t(x_t)).
\end{align}
Using this relation, the fake-teacher score difference can be written as:
\begin{align}
    s_\mathrm{fake}(\hat x_t; t) - s_\mathrm{real}(\hat x_t; t) = t^{-1}(1-t)[v_\mathrm{real}(\hat x_t; t) - v_\mathrm{fake}(\hat x_t; t)].
\end{align}
By letting $\gamma_t = t^{-1}(1-t)$, the DMD objective coincides with
\begin{align}
    \nabla_\theta\mathcal L_\mathrm{DMD} = \mathbb E\left[\gamma_t(v_\mathrm{real}(\hat x_t; t) - v_\mathrm{fake}(\hat x_t; t))\frac{d\hat x_t}{d\theta}\right].
\end{align}

\subsection{MeanFlow as a Flow Map}

Let the ground-truth flow map be $X(x_t; t, s) = x_t + \int^s_t v^*_\tau(x_\tau)d\tau$. We assume that a flow-map network $X_\theta$ approximates this map, $X_\theta(x_t; t, s)\approx X(x_t; t, s)$. Under the MeanFlow-style realization, $X_\theta(x_t; t, s) = x_t + (s-t)u_\theta(x_t; t, s)$, where $u_\theta$ is a pseudo-velocity network. Consistency distillation achieves this approximation through the following objective:
\begin{align}
    \nabla_\theta\mathcal L_\mathrm{CD} &= \nabla_\theta\mathbb E_{x\sim p_\mathrm{real}, z\sim p_Z, t,s\sim p_T}\left[\tilde w_{t,s}X_\theta(x_t; t, s)\cdot\operatorname{sg}\left[\frac{dX_\theta(x_t; t, s)}{dt}\right]\right] \\
    & = \mathbb E\left[\tilde w_{t,s}(s-t)\nabla_\theta u_\theta(x_t; t, s)\cdot\operatorname{sg}\left[v^*_t - u_\theta(x_t; t, s) + (s-t)\frac{d}{dt}u_\theta(x_t; t, s)\right]\right] \\
    &\begin{aligned}\ = \nabla_\theta\mathbb E&\left[\tilde w_{t,s}\frac{t-s}2\Big\|-u_\theta(x_t; t, s) + \operatorname{sg}\Big[\right.\\
        &\left.\left.\left.u_\theta(x_t; t, s) + v^*_t - u_\theta(x_t; t, s) + (s-t)\frac{d}{dt}u_\theta(x_t; t, s)\right] \right\|^2_2\right]
    \end{aligned} \\
    &= \nabla_\theta\mathbb E\left[\tilde w_{t,s}\frac{t-s}{2}\left\|u_\theta(x_t; t, s) - \operatorname{sg}\left[v^*_t + (s-t)\frac{d}{dt}u_\theta(x_t; t, s)\right]\right\|^2_2\right],
\end{align}
where $v^*_t = v^*_t(x_t)$ and $\tilde w_{t,s}$ is a scalar loss weight. This coincides with the MeanFlow identity when $\tilde w_{t,s} = 2/(t-s)$ and with \cref{eq:ct} when $w_{t,s} = \tilde w_{t,s}(t-s)/2$. This explains why MeanFlow can be interpreted as a consistency model and, further, as a flow-map model~\citep{boffi2025flow}.

\subsection{Deriving FSF-DMD}
\label[appendix]{app:deriving-fsf-dmd}

Based on this setting, let the generator be $G_\theta(z) = X_\theta(z; 1, 0)$. The DMD objective is
\begin{align}
    \nabla_\theta\mathcal L_\mathrm{DMD} = \mathbb E_{z,z'\sim p_Z,t\sim p_T}\left[\gamma_t(v_\mathrm{real}(\hat x_t; t) - v_\mathrm{fake}(\hat x_t; t))\frac{d\hat x_t}{d\theta}\right],
\end{align}
where $\hat x = G_\theta(z)$ and $\hat x_t = (1-t)\hat x + tz'$. Typical DMD trains an auxiliary fake-velocity network $v_\psi$ to match the ground-truth fake velocity $v_\mathrm{fake}$:
\begin{align}
    \mathcal L_\psi = \mathbb E_{z,z',t}\left[\left\|v_\psi(\hat x_t; t) - v_t(\hat x_t|\hat x) \right\|^2_2\right],\quad v_\psi(\hat x_t; t)\approx v_\mathrm{fake}(\hat x_t; t)= \mathbb E_{\hat x|\hat x_t}[v_t(\hat x_t|\hat x)],
\end{align}
where $v_t(\hat x_t|\hat x) = t^{-1}(\hat x_t - \hat x) = z' - \hat x$. In general, this conditional expectation over the posterior $p(\hat x|\hat x_t)$ is nontrivial. We note that changing the noising process for $\hat x_t$ alters the posterior distribution, and the target fake velocity $v_\mathrm{fake}$ also changes. DMD2 assumes the independent coupling, $\hat x \perp z'$ for $z'\sim p_Z$, so $z\ne z'$ in general.

\textbf{Motivating Observation.} Since $\hat x = G_\theta(z) = z + (0 - 1)u_\theta(z; 1, 0) = z - u_\theta(z; 1, 0)$, assuming $z = z'$ gives $v_t(\hat x_t|\hat x) = z - \hat x = u_\theta(z; 1, 0)$. From this perspective, the fake velocity under the prior-space posterior becomes $\mathbb E_{z|\hat x_t}[u_\theta(z; 1, 0)]$. This observation tells us that the conditional fake velocity can be written using the pseudo-velocity $u_\theta$. However, since the posterior $p(z|\hat x_t)$ is nontrivial, the conditional expectation does not reduce to analytic terms that we can compute practically. Thus, we explore couplings under which the conditional expectation becomes analytically accessible.

\textbf{Exploring the Coupling.} For the independent coupling, $\hat x\perp z'$, the fake velocity becomes $v_\mathrm{fake}(\hat x_t; t) = (\hat x_t - \mu_t^\mathrm{IC}(\hat x_t))/t$ for $\mu^\mathrm{IC}_t(\hat x_t) = \mathbb E_{\hat x|\hat x_t}[\hat x]$. In this case, the conditional posterior is $p(z|\hat x_t) \propto p_Z(z)p(\hat x_t|z)$, and the generative posterior mean $\mu^\mathrm{IC}_t(\hat x_t)$ follows
\begin{align}
    \mu_t^\mathrm{IC}(\hat x_t) = \mathbb E_{z\sim p_Z}\left[\frac1{Z^\mathrm{IC}(\hat x_t)}\exp\left(-\frac1{2t^2}\|\hat x_t - (1-t)G(z)\|^2_2\right)\cdot G(z)\right]
\end{align}
for the partition function $Z^\mathrm{IC}(\hat x_t)$. This is nontrivial since it requires an expectation over the support of the prior distribution $p_Z$.

For the generator-induced coupling considered above, $z = z'$, which gives $\hat x_t = I_t(z) = (1-t)G(z) + tz$. The posterior mean $\mu^\mathrm{GIC}_t(\hat x_t)$ is given by
\begin{align}
    \mu^\mathrm{GIC}_t(\hat x_t) = \mathbb E_{z\sim p_Z}\left[\frac1{Z^\mathrm{GIC}(\hat x_t)}\delta(I_t(z) - \hat x_t)\cdot G(z)\right]
\end{align}
for the corresponding partition function $Z^\mathrm{GIC}(\hat x_t)$ and the delta function $\delta$. Since there may be multiple $z$ satisfying $I_t(z) = \hat x_t$, this is also nontrivial, and the corresponding fake velocity $v_\mathrm{fake}^\mathrm{GIC}(\hat x_t; t) = (\hat x_t - \mu_t^\mathrm{GIC}(\hat x_t)) / t$ is not directly accessible either.

Lastly, consider the PF-ODE-based coupling where the noised sample $\hat x_t$ is given by the flow map $\hat x_t = X(\hat x; 0, t)$. When the marginal velocity is Lipschitz continuous, the solution of the ODE $dx_t = v^*_t(x_t)dt$ uniquely exists by the Picard-Lindel\"of theorem. The flow mapping is therefore injective and invertible in time, $X(X(x_t; t, s); s, t) = x_t$.

Based on this property, $\hat x_t$ is uniquely determined by the given $\hat x$, so the posterior distribution collapses to the point mass $p(\hat x|\hat x_t) = \delta(X(\hat x; 0, t) - \hat x_t)$. By injectivity and invertibility, $p(\hat x|\hat x_t) = \delta(\hat x - X(\hat x_t; t, 0))$, and the posterior mean $\mu_t^\mathrm{ODE}(\hat x_t)$ becomes accessible:
\begin{align}
    \mu_t^\mathrm{ODE}(\hat x_t) = \mathbb E_{\hat x|\hat x_t}[\hat x] = \mathbb E_{\delta(\hat x - X(\hat x_t; t, 0))}[\hat x] = X(\hat x_t; t, 0).
\end{align}

With this posterior mean, the fake velocity is
\begin{align}
    v_\mathrm{fake}^\mathrm{ODE}(\hat x_t; t) = \frac{\hat x_t - \mu_t^\mathrm{ODE}(\hat x_t)}{t} = \frac{\hat x_t - X(\hat x_t; t, 0)}{t}.
\end{align}
Substituting $X$ with the approximating network $X_\theta$ under the MeanFlow-style realization gives
\begin{align}
    v_\mathrm{fake}^\mathrm{ODE}(\hat x_t; t)\approx \frac{\hat x_t - X_\theta(\hat x_t; t, 0)}{t} = \frac{\cancel{\hat x_t - \hat x_t} + tu_\theta(\hat x_t; t, 0)}{t} = u_\theta(\hat x_t; t, 0).
\end{align}
Thus, we obtain the objective in \cref{eq:fsf-dmd-ideal}, which does not rely on an explicit fake-velocity network:
\begin{align}
    \Delta_{u, \Phi} = u_\theta(\hat x_t; t, 0) - v_\Phi(\hat x_t; t),\quad\nabla_\theta\mathcal L(u, \Phi) = \mathbb E\left[-\gamma_t\Delta_{u, \Phi}\frac{d\hat x_t}{d\theta}\right],
\end{align}
for the teacher-velocity network $v_\Phi(x_t; t)\approx v^*_t(x_t)$.

In summary, (i) PF-ODE-based coupling gives a point-mass posterior, and (ii) the fake velocity reduces to the endpoint pseudo-velocity $u_\theta(\hat x_t; t, 0)$ under the MeanFlow-style generator formulation.

This derivation assumes Lipschitz continuity of the marginal velocity, PF-ODE-based coupling, and a flow-map-based generator. The first is a common assumption in the flow-matching literature, and the last has been explored recently in \cite{cheng2026twinflow, ge2026senseflow, zheng2026large}.

We use $\hat x_t = (1-t)\hat x + tz'$ for the DMD objective following previous work. The gradient path then becomes simpler, $d\hat x_t/d\theta = (1-t)d\hat x/d\theta$, and the final objective is
\begin{align}
    \nabla_\theta \mathcal L_\mathrm{FSF\text{-}DMD}(u, \Phi) &= \mathbb E\left[-\tilde w_t\gamma_t\Delta_{u, \Phi}\frac{d\hat x_t}{d\theta}\right] = \mathbb E\left[-\tilde w_t\gamma_t(1-t)\Delta_{u, \Phi}\frac{d\hat x}{d\theta}\right] \\
    &= \mathbb E\left[\tilde w_t\gamma_t(1-t)\Delta_{u, \Phi}\frac{du_\theta(z; 1, 0)}{d\theta}\right] \\
    &= \nabla_\theta\mathbb E\left[\frac12\tilde w_t\gamma_t(1-t)\left\|u_\theta(z; 1, 0) - \operatorname{sg}\left[u_\theta(z; 1, 0) - \Delta_{u, \Phi}\right]\right\|^2_2\right]
\end{align}
since $\hat x = X_\theta(z; 1, 0) = z - u_\theta(z; 1, 0)$ and $d\hat x/d\theta = -du_\theta(z; 1, 0)/d\theta$. Setting $w_t = \frac12\tilde w_t\gamma_t(1-t)$ for the scalar weight $\tilde w_t\ge0$ gives the objective in \cref{eq:fsf-dmd-impl}. The training algorithm follows \cref{alg:fsf-dmd-training}.

\begin{algorithm}[th]
    \caption{FSF-DMD distillation training}
    \label{alg:fsf-dmd-training}
    \begin{algorithmic}[1]
        \State {\bfseries Input:} Noise distribution $p_Z$, labeled data distribution $p_\mathrm{real}$, time distribution $p_T$, number of backward-simulation steps $M$, loss weights $w_{t,s}$ and $w_t$, DMD coefficient $\lambda$, learning rate $\eta$, CFG scale $\omega$, timestep-shifting parameter $\kappa$.
        \Repeat
            \State $z \sim p_Z,\quad (x,c) \sim p_\mathrm{real}$
            \State $t, s\sim p_T,\quad t, s\gets \max\{t, s\}, \min\{t, s\}$
            \State $x_t\gets (1-t)x + tz$
            \State $v_t \gets \omega v_\Phi(x_t; t, c) + (1-\omega)v_\Phi(x_t; t, \varnothing)$ \Comment{Classifier-free guidance}
            \State $u_{t,s}, u'_{t,s}\gets \mathrm{JVP}(u_\theta, (x_t, t, s), (v_t, 1, 0))$
            \State $\mathcal L_\mathrm{CD}\gets w_{t,s}\|u_{t,s}- \operatorname{sg}[v_t + (s - t)u_{t,s}']\|^2_2$ \Comment{Consistency distillation in \cref{eq:cd}}
            \State
            \State $\bar u \gets 0$ \Comment{Backward simulation in \cref{eq:tilde-F}}
            \For{$i \gets M$ to $1$}
                \State $\hat x_{i/M} \gets z - \bar u$
                \State $\bar u\gets \bar u + u_\theta(\operatorname{sg}[\hat x_{i/M}]; i / M, (i - 1) / M) / M$
            \EndFor
            \State $\tau\sim \mathcal U[0, 1),\quad t\gets \kappa \tau/((\kappa - 1)\tau + 1)$ \Comment{Timestep shifting}
            \State $z'\sim p_Z$
            \State $\hat x \gets z - \bar u,\quad \hat x_t \gets (1-t)\hat x + tz'$
            \State $\hat v_t \gets \omega v_\Phi(\hat x_t; t, c) + (1 - \omega)v_\Phi(\hat x_t; t, \varnothing)$ \Comment{Classifier-free guidance}
            \State $\Delta_{u, \Phi}\gets u_\theta(\hat x_t; t, 0) - \hat v_t$
            \State $\mathcal L_{\bar u, \Phi}\gets w_t\|\bar u - \operatorname{sg}[\bar u - \Delta_{u, \Phi}]\|^2_2$ \Comment{DMD surrogate in \cref{eq:bwd-simulation}}
            \State $\mathcal L_\mathrm{distill} \gets \mathcal L_\mathrm{CD} + \lambda\mathcal L_{\bar u,\Phi}$ \Comment{Full objective in \cref{eq:fsf-dmd-distill}}

            \State
            \State $\theta\gets \theta-\eta\nabla_\theta \mathcal L_\mathrm{distill}$ \Comment{Update parameters}
        \Until convergence
    \end{algorithmic}
\end{algorithm}

\subsection{Considering PF-ODE-Based Noising in Practice}
\label[appendix]{app:consider-pf-ode}

Since we alter the perturbation process, the resulting fake-velocity field can be inconsistent with the teacher-velocity field. To handle this, we considered using PF-ODE noising in practice and identified several concerns: (i) It would increase optimization complexity. Since the exact flow map is intractable, we would need to approximate the noised sample as $\hat x_t = X_\theta(\hat x; 0, t)$. As flow-map models empirically assume a mapping toward the data distribution, learning a mapping toward the prior distribution may impose an additional burden on the generator. Additionally, backpropagation through $d\hat x_t/d\hat x$ involves the Jacobian $\nabla X_\theta(\hat x; 0, t)$, while independent perturbation gives the simpler form $d\hat x_t/d\hat x = 1-t$. This is computationally inefficient and may increase optimization complexity.

(ii) Using PF-ODE noising while keeping the teacher unchanged would introduce an additional inconsistency with the teacher-velocity field. Alternatively, adapting the teacher network to the same noising field would make its velocity coincide with the endpoint pseudo-velocity by construction. The correction term would then collapse to zero, eliminating its effect.

We therefore retain independent perturbation as a practical engineering choice.

\subsection{Related Work}
\label[appendix]{app:related-work}

\textbf{DMD.} The implementation-level objective of DMD is
\begin{align}
    \mathcal L_\mathrm{DMD} = \frac12\cdot \mathbb E_{z,z',t}\left[\left\|G_\theta(z) - \operatorname{sg}\left[G_\theta(z) - \frac{\mu_\psi(\hat x_t; t) - \mu_\Phi(\hat x_t; t)}{|G_\theta(z) - \mu_\Phi(\hat x_t; t)|}\right]\right\|^2_2\right].
\end{align}
In the velocity space with $G_\theta(z) = z - u_\theta(z; 1, 0)$, we use the following objective to train DMD2:
\begin{align}\label{eq:dmd-v}
    \mathcal L^v_\mathrm{DMD} = \frac12 \cdot \mathbb E_{z,z',t}\left[\left\|u_\theta(z; 1, 0) - \operatorname{sg}\left[u_\theta(z; 1, 0) - \frac{v_\psi(\hat x_t; t) - v_\Phi(\hat x_t; t)}{| u_\theta(z; 1, 0) - v_\Phi(\hat x_t; t)|}\right]\right\|^2_2\right].
\end{align}
The fake-velocity network is trained with flow matching:
\begin{align}\label{eq:dmd-fake-velocity}
    \mathcal L_\psi = \mathbb E_{z,z',t}\left[\left\|v_\psi(\hat x_t; t) - (z' - \hat x)\right\|^2_2\right],\quad v_\psi(\hat x_t; t)\approx v_\mathrm{fake}(\hat x_t; t).
\end{align}
For the auxiliary GAN loss, the discriminator is trained with
\begin{align}
    \mathcal L_D = \mathbb E[\sigma_+(-r_\mathrm{real}) + \sigma_+(r_\mathrm{fake})],\quad \mathcal L_G = \mathbb E[\sigma_+(-r_\mathrm{fake})],
\end{align}
where $r_\mathrm{real}$ and $r_\mathrm{fake}$ are the corresponding discriminator logits, and $\sigma_+$ is the softplus function. Under TTUR, the discriminator objective $\mathcal L_D$ is optimized during fake-side updates, $\mathcal L_\mathrm{fake}$, and $\mathcal L_G$ is optimized during generator updates, $\mathcal L_\mathrm{gen}$:
\begin{align}\label{eq:dmd2-objective}
    \mathcal L_\mathrm{fake} = \mathcal L_\psi + \lambda_D \mathcal L_D,\quad \mathcal L_\mathrm{gen} = \mathcal L^v_\mathrm{DMD} + \lambda_G\mathcal L_G.
\end{align}

\textbf{rCM}~\citep{zheng2026large} is based on sCD~\citep{lu2025simplifying} and adopts distribution matching to regularize flow-map models toward a self-generated correction. It assumes the trigonometric interpolation $x_t = \cos(t) x + \sin(t)z$ for $t\in [0, \pi/2]$ and trains a model with tangent normalization:
\begin{align}
    \mathcal L_\mathrm{sCD} = \mathbb E_{x,z,t}\left[\left\|u_\theta(x_t; t, 0) - \operatorname{sg}\left[u_\theta(x_t; t, 0) + \frac{g}{\|g\|_2 + c}\right]\right\|^2_2\right],\quad g = w_t \frac{dX_\theta(x_t; t, 0)}{dt}, \\
    \frac{dX_\theta(x_t; t, 0)}{dt} = -\cos(t) (u_\theta(x_t; t, 0) - v_\Phi(x_t; t)) - \sin(t)\left(x_t + \frac{du_\theta(x_t; t, 0)}{dt}\right),
\end{align}
where $w_t = \cos(t)$ and $c = 0.1$ in this case. With a MeanFlow-style realization under linear interpolation and $s$ relaxed from the fixed constant zero, this becomes
\begin{align}
    \mathbb E_{x,z,t,s}\left[\left\|u_\theta(x_t; t, s) - \operatorname{sg}\left[u_\theta(x_t; t, s) + \frac{g}{\|g\|_2 + c}\right]\right\|^2_2\right], \\
    g = v_\Phi(x_t; t) - u_\theta(x_t; t, s) + (s-t)\frac{d}{dt}u_\theta(x_t; t, s),
\end{align}
when $w_t = 1$. This objective coincides with \cref{eq:cd} when the tangent normalization $1 / (\|g\|_2 + c)$ is discarded. We found that \cref{eq:cd} can be stably trained without tangent normalization, and adding it degrades the consistency distillation performance in our setup. Therefore, we do not use it for this case, and jointly train the DMD objective with the consistency distillation objective in \cref{eq:cd}:
\begin{align}\label{eq:rcm-objective}
    \mathcal L_\mathrm{rCM}^v = \mathcal L_\mathrm{CD} + \lambda_\mathrm{rCM} \mathcal L^v_\mathrm{DMD},
\end{align}
where $\lambda_\mathrm{rCM} = 0.01$ following the original work. For backward simulation, rCM first samples $j\sim \mathcal U[0, N)$ in \cref{eq:dmd2-bwd-simulation} and constructs $\hat x$ as $\hat x = G_\theta(\operatorname{sg}[\hat x_{t_{j'}}^{(j')}]; t_{j'})$ for $j' = N - j$. We leave the comparison with our formulation to future work.

\textbf{SenseFlow}~\citep{ge2026senseflow} introduces Implicit Distribution Alignment (IDA) and Intra-Segment Guidance (ISG). To make the fake-score network track the generator more closely, it introduces IDA, which explicitly adds a parameter-blending stage after the generator updates:
\begin{align}\label{eq:ida}
    \psi\gets\lambda_\mathrm{IDA}\psi + (1 - \lambda_\mathrm{IDA})\theta.
\end{align}
In this comparison, we evaluate IDA to compare design choices of the fake-side tracker in our setting. Following the official implementation, we use $\lambda_\mathrm{IDA} = 0.97$ by default.

Since the generation timesteps of DMD2 are uniformly determined, the timesteps between adjacent steps are not provided to the generator. To regularize the generator over these segments, SenseFlow introduces the ISG objective, $\mathcal L_\mathrm{ISG}$:
\begin{align}
    \mathcal L_\mathrm{ISG} &= \mathbb E_{i\sim\mathcal U[1, N], \tau\sim \mathcal U[t_{i-1}, t_i)}\left[\left\|X_\theta(\hat x_{t_i}; t_i, t_{i-1}) - \operatorname{sg}\left[\hat x_\mathrm{tar}\right]\right\|^2_2\right],\\
    \hat x_\mathrm{tar} &= X_\theta(\hat x_{t_i} + (\tau - t_i) v_\Phi(\hat x_{t_i}; t_i); \tau, t_{i-1}).
\end{align}
This can be interpreted as an RCGM~\citep{sun2026anystep}-style flow map on generative distributions. Since we already use RCGM when training from scratch, we replace the ISG evaluation with RCGM. We leave the discriminator based on visual foundation models in SenseFlow to future work.

\textbf{TwinFlow.} To avoid an additional fake-side tracker, TwinFlow trains the pseudo-velocity network to estimate the fake velocity under an augmented condition:
\begin{align}
    \mathcal L_\mathrm{adv} = \mathbb E_{z,z',t}\left[\left\|u_\theta(\hat x_t; -t, -t) - (z' - \hat x)\right\|^2_2\right],\quad u_\theta(\hat x_t; -t, -t)\approx v_\mathrm{fake}(\hat x_t; t).
\end{align}
Distribution matching is then performed using this fake-velocity estimate:
\begin{align}\label{eq:twin-dm}
    \mathcal L_\mathrm{DM} &= \frac12\mathbb E_{z,z',t}\left[\left\|u_\theta(z; 1, 0) - \operatorname{sg}\left[u_\theta(z; 1, 0) - \mathrm{clip}(\Delta_\mathrm{Twin}, -1, 1)\right]\right\|^2_2\right], \\
    \Delta_\mathrm{Twin} &= u_\theta(\hat x_t; -t, -t) - v_\Phi(\hat x_t; t).
\end{align}
For the training-from-scratch setting, TwinFlow approximates the teacher velocity as $v_\Phi(\hat x_t; t)\approx u_\theta(\hat x_t; t, t)$, using the timestep-limit property $s\to t$.

We use improved MeanFlow as the flow-map baseline, whereas TwinFlow uses RCGM:
\begin{align}
    \mathcal L_\mathrm{RCGM} &= \mathbb E_{x,z,t,s}\left[\left\|u_\theta(x_t; t, s) - \operatorname{sg}\left[u_\theta(x_t; t, s) - \mathrm{clip}\left(\Delta_\mathrm{RCGM}, -1, 1\right)\right]\right\|^2_2\right], \\
    \Delta_\mathrm{RCGM} &= (t-s)\epsilon^{-1}\cdot(u_\theta(x_t; t, s) - u_\mathrm{tar}),\\
    u_\mathrm{tar} &= \frac{x_t - X_\theta(X_\theta(x_t - \epsilon\cdot v_t; t - \epsilon, t_\mathrm{mid}); t_\mathrm{mid}, s)}{t - s},\quad t_\mathrm{mid} = \frac{t - \epsilon + s}{2},
\end{align}
where a two-step process is assumed. For comparison, we evaluate TwinFlow with both base flow-map objectives in the training-from-scratch setting.

\section{Experimental Details}
\label[appendix]{app:experimental-details}

\subsection{Implementation}

\begin{table}[th]
\centering
\caption{Experimental settings.}
\label{tab:experimental-setting}
\vspace{1em}
\resizebox{1.0\linewidth}{!}{%
\begin{tabular}{lccc}
\toprule
Dataset & \multicolumn{3}{c}{ImageNet-1K $256\!\times\!256$} \\
\midrule
Preprocessor & \multicolumn{3}{c}{VA-VAE~\citep{yao2025vavae}} \\
Input size & \multicolumn{3}{c}{$16\times 16\times 32$} \\
Condition & \multicolumn{3}{c}{Conditional} \\
\midrule
Backbone & LightningDiT-B/1$^\dagger$~\citep{yao2025vavae} & LightningDiT-B/1 & LightningDiT-B/1$^\dagger$ \\
Teacher & LightningDiT-B/1$^\dagger$ & LightningDiT-B/1 & - \\
Initialization & iMF-B/1~\citep{geng2025improvedmeanflowschallenges} & LightningDiT-B/1 & Random initialization \\
\midrule
Params (M) & 134 & 131 & 134 \\
Depth & 12 & 12 & 12\\
Hidden dim & 768 & 768 & 768 \\
Heads & 12 & 12 & 12 \\
Patch size & $1\!\times\!1$ & $1\!\times\!1$ & $1\!\times\!1$ \\
\midrule
Loss & $\mathcal L_\mathrm{distill}$ (Eq.~\ref{eq:fsf-dmd-distill}) & $\mathcal L_\mathrm{distill}$ & $\mathcal L_\mathrm{scratch}$ (Eq.~\ref{eq:fsf-dmd-scratch})\\
JVP & Approx. ($\epsilon=0.005$) & Approx. ($\epsilon=0.005$) & Approx. ($\epsilon=0.005$) \\
$w_{t,s}$ & $\cos(\pi t/2)$ & $\cos(\pi t/2)$ & $\cos(\pi t/2)$ \\
$w_t$ & $1/|\bar u_\theta(z; 1, 0) - v_\Phi(\hat x_t; t)|$ & $\cos(\pi t/2)/|\bar u_\theta(z; 1, 0) - v_\Phi(\hat x_t; t)|$ & $\cos(\pi t/2)$ \\
$\lambda$ & 0.1 & 0.001 & 0.01 \\
$\kappa$ & 10.0 & 10.0 & 1.0 \\
$M$ & 2 & 2 & 1 \\
CFG scale $\omega$ & 6.0 & 6.0 & 6.0 \\
CFG range & [0.0, 0.9) & [0.0, 0.9) & [0.0, 0.9)\\
Training steps & 4K & 20K + 8K & 400K \\
\midrule
FP precision & \multicolumn{3}{c}{BF16} \\
Batch size & \multicolumn{3}{c}{128} \\
Label dropout & - & - & 0.1 \\
Optimizer & \multicolumn{3}{c}{NorMuon~\citep{li2025normuonmakingmuonefficient}} \\
LR scheduler & \multicolumn{3}{c}{Constant} \\
$\beta_1$ & \multicolumn{3}{c}{-} \\
$\beta_2$ & \multicolumn{3}{c}{0.95} \\
Momentum & \multicolumn{3}{c}{0.95} \\
Weight decay & \multicolumn{3}{c}{0} \\
Learning rate & \multicolumn{3}{c}{$5\times 10^{-4}$} \\
EMA decay & \multicolumn{3}{c}{0.99995}\\
\bottomrule
\end{tabular}
}
\end{table}

\textbf{Compute Resources.} Our distillation experiments are run on a GPU cloud server with a 72-core CPU, 2TB RAM, 5.1TB storage, and four NVIDIA B200 GPUs. Each reported run uses a single B200 GPU. We report wall-clock seconds per training step in \cref{tab:additional-metrics}.

The training-from-scratch experiments are run on a GPU cloud server with a 128-core CPU, 2TB RAM, 26TB storage, and eight NVIDIA A100 GPUs. Each reported run uses a single A100 GPU. The wall-clock seconds per training step are: iMF baseline 0.78 sec/step, iMF + DMD2 1.88 sec/step, iMF + TwinFlow 1.03 sec/step, RCGM 1.46 sec/step, RCGM + TwinFlow 1.67 sec/step, and FSF-DMD 0.86 sec/step. Combining the reported training steps with the measured sec/step values gives an approximate per-run GPU-hour estimate.

\textbf{ImageNet-1K.} For ImageNet-1K, we evaluate experiments at $256\!\times\!256$ resolution and encode images into $16\!\times\!16\!\times\!32$ latent representations with VA-VAE~\citep{yao2025vavae}. We use LightningDiT~\citep{yao2025vavae} as the network backbone throughout, with RMSNorm, QK-normalization, and RoPE enabled, following \cite{li2026back}. We use a fixed base-size transformer with $1\!\times\!1$ patch size, denoted by B/1. Since we use NorMuon~\citep{li2025normuonmakingmuonefficient} by default, we use an Adam optimizer~\citep{2015-kingma} for non-matrix weights with a learning rate of $10^{-4}$, momentum parameters $\beta = (0.9, 0.95)$, and a maximum gradient norm of 0.1.

\textbf{Input Augmentation for iMF.} To make the network compatible with improved MeanFlow (iMF)~\citep{geng2025improvedmeanflowschallenges}, we augment the network, denoted by $\dagger$, to accept the classifier-free guidance (CFG)~\citep{ho2021classifierfree} scale $\omega$ and CFG range $t_\mathrm{min}, t_\mathrm{max}$. Following iMF, each input except the sample $x_t$ is embedded into a token and repeated $N_\mathrm{repeat}$ times with a learnable positional embedding. These tokens are used for adaptive layer normalization and are concatenated to the sample tokens after $N_\mathrm{icl}$ transformer layers, jointly passed through the remaining transformer layers~\citep{li2026back}. We use $N_\mathrm{repeat} = 8$ for the class label, $N_\mathrm{repeat} = 4$ for other tokens, and $N_\mathrm{icl}=4$ by default. For consistency training and distillation, $\omega$ is sampled from a log-uniform distribution over $\omega\in[1.0, 8.0)$, with $t_\mathrm{min}\sim \mathcal U[0, 0.5)$ and $t_\mathrm{max}\sim \mathcal U[0.5, 1.0)$ when using this augmented backbone. We set $\omega=1$ when $t\notin[t_\mathrm{min}, t_\mathrm{max})$.

\textbf{JVP Operation.} We use the finite-difference approximation for the JVP operation~\citep{Wang_2026_CVPR, zheng2026large},
$\mathrm{JVP}(u_\theta, (x_t, t, s), (v_t, 1, 0)) = (u_\theta(x_t + \epsilon\cdot v_t; t + \epsilon, s) - u_\theta(x_t - \epsilon\cdot v_t; t - \epsilon, s)) / (2\epsilon)$, wherever a JVP is required.

\textbf{Flow-Map Training.} For consistency training and consistency distillation, $t$ and $s$ are sampled from $p_T=\mathrm{Beta}(0.8, 1.0)$ and then reordered as $t, s\gets \max\{t, s\}, \min\{t, s\}$ to define maps toward higher-SNR regions~\citep{cheng2026twinflow}. We randomly drop 50\% of $s$ to $t$ to anchor the training signal to flow matching~\citep{geng2026mean}, and additionally drop 10\% of class labels to the null label $\varnothing$ for unconditional generation in consistency training. Following the v-loss perspective of \cite{geng2025improvedmeanflowschallenges}, we replace $v_\Phi(x_t; t, c)$ in the matching target with $v_t(x_t|x)$ while retaining $v_\Phi(x_t; t, c)$ in the JVP. We use $w_{t,s} = \cos(\pi t/2)$ by default and found that combining the cosine similarity loss of \cite{yao2025vavae} with the pseudo-Huber loss, $\sqrt{\mathcal L + \nu^2} - \nu$ for $\nu=5$, improves performance.

Unlike the augmented network setting, general flow-matching networks take samples $x_t$ and current timesteps $t$ as input, and do not assume target timesteps $s$. To pass target timesteps to the flow-matching-initialized generator, we use the conditioning scheme of \cite{lee2026decoupled}, which conditions the initial transformer layers on $t$ and the remaining transformer layers on $s$. For the base-size transformer, we pass $t$ to the first eight layers and $s$ to the remaining layers following the official implementation.

We note that the official implementation of LightningDiT~\citep{yao2025vavae} uses 3-channel CFG, which applies CFG to the first 3 channels of the latent representation and retains the remaining 29 channels as the class-conditional velocity. We observe that this 3-channel CFG degrades the FID to 14.99 for the augmented backbone setting. It is beneficial for the original LightningDiT-B/1 setting, improving FSF-DMD from 4.96 to 4.70 FID. The DMD2 baseline shows the same trend, 7.92 to 5.53 FID, and we use 3-channel CFG for this setting.

\textbf{Distribution Matching.} For distribution matching, we sample $t$ from a uniform distribution and optionally shift it using $t \gets \kappa t/((\kappa - 1)t + 1)$ with timestep-shifting parameter $\kappa$. We clamp $t$ to $t_\mathrm{max} - 10^{-3}$ to pass the unmasked $\omega$. We found that timestep shifting underperforms the uniform distribution for DMD2, and therefore keep the uniform distribution from previous work in this case. We fix the CFG parameters without sampling, even in the augmented setting, and use the EMA network for the fake-side surrogate. In the flow-matching-initialized setting, we warm up $\lambda$ for 20K training steps and then train the full objective for another 8K steps. For a fair comparison, we additionally train DMD2 on the consistency-distilled networks and report the better FIDs in \cref{tab:ldit-teacher}. We adopt backward simulation, which generates multistep samples and then trains the network with perturbed ones. In the training-from-scratch setting, we found that backward simulation underperforms single-step generation for distribution matching. We therefore report simulation steps separately from the generation steps used for inference, and use a single-step simulation for the training-from-scratch setting.

For DMD2, we train the fake-velocity network with \cref{eq:dmd-fake-velocity} and use the objective in \cref{eq:dmd-v} for distribution matching. We explore TTUR ratios from 1:1 to 1:5, training the fake-velocity network for up to five steps per generator update. For the GAN loss, previous work uses the bottleneck layer of the U-Net architecture to construct a discriminator branch. Since DiT does not include a bottleneck layer, we use DDT~\citep{wang2025ddtdecoupleddiffusiontransformer} for the real/fake discriminator branch. We branch from the fifth layer of the fake-velocity network and add a two-block decoupled network with a concatenated class token. We pass the class-token output to a projection layer to obtain the discriminator logit $r$. The discriminator is trained with \cref{eq:dmd2-objective} using $\lambda_G = \lambda_D = 5\times 10^{-3}$, following the official implementation.

We observed that TwinFlow underperforms compared to the baseline methods. One possible explanation for this behavior in both settings is that TwinFlow adds the auxiliary objective $\mathbb E\|u_\theta(\hat x_t; -t, -t) - (z' - \hat x)\|^2_2$ on top of the original flow-map objective. This requires the same network to jointly optimize the generative task and the additional fake-side estimation task, potentially introducing extra optimization and representational burden. The degree of this interference may depend on the underlying flow-map objective, which could explain why the degradation is more pronounced for RCGM than for iMF.

We compare our method with design choices adopted in prior work. We set the baselines to consistency distillation (AYF-EMD)~\citep{sabour2026align} without adversarial fine-tuning and DMD2~\citep{yin2024improved}. We then explore several DiT-adapted variants, including joint training with consistency distillation~\citep{zheng2026large} in \cref{eq:rcm-objective}, IDA~\citep{ge2026senseflow} in \cref{eq:ida}, and the TwinFlow-style variants~\citep{cheng2026twinflow} described in \cref{eq:twin-dm}. Network backbones, basic objectives, CFG parameters, optimizers, and maximum training steps are kept fixed, while the objective and relevant hyperparameters are modified for controlled comparison.

The detailed parameters are listed in \cref{tab:experimental-setting}.

\subsection{Additional Experiments}
\label[appendix]{app:ablation-study}

\begin{figure}[th]
\begin{minipage}{1.0\textwidth}
\centering
\begin{minipage}{0.4\textwidth}
\begin{minipage}[th]{1.0\linewidth}
    \centering
    \captionof{table}{Results of FSF-DMD ($M=2$) across $\kappa$ when $\lambda\times w_t=0.05\times 1.0$.}
    \label{tab:ab-gamma}
    \resizebox{1.0\linewidth}{!}{%
    \begin{tabular}{c|ccccc}
        \toprule
        $\kappa$ & 1.0 & 3.0 & 5.0 & 7.0 & 10.0 \\
        \midrule
        FID & 4.06 & 3.94 & 3.89 & \textbf{3.87} & \textbf{3.87} \\
        \bottomrule
    \end{tabular}
    }
\end{minipage}\\
\begin{minipage}[th]{1.0\linewidth}
    \vspace{1em}
    \centering
    \captionof{table}{FIDs across $\lambda$ and $w_t$. Numeric entries in the header row are $\lambda$ ($\kappa=10$).}
    \label{tab:ab-lambda}
    \resizebox{1.0\linewidth}{!}{%
    \begin{tabular}{c|cccc}
        \toprule
        $w_t$ & 0.01 & 0.05 & 0.1 & 0.5 \\
        \midrule
        1 & 3.97 & 3.87 & 3.98 & 3.99 \\
        $1/|\bar u_\theta - v_\Phi|$ & 4.03 & \textbf{3.85} & \textbf{3.85} & 3.94 \\
        \bottomrule
    \end{tabular}
    }
\end{minipage}
\end{minipage}\hspace*{1em}
\begin{minipage}{0.5\textwidth}
    \centering
    \captionof{table}{Results of DMD2 across TTURs.}
    \label{tab:dmd2-sweep}
    \resizebox{1.0\linewidth}{!}{%
    \begin{tabular}{lcc}
        \toprule
        DMD2 variants & FID w/o GAN & FID w/ GAN \\
        \midrule
        TTUR=1:1 & 4.38 & 4.28 \\
        TTUR=1:2 & 4.32 & 4.27 \\
        TTUR=1:3 & 4.35 & 4.25 \\
        TTUR=1:4 & 4.35 & 4.30 \\
        TTUR=1:5 & \textbf{4.23} & \textbf{4.18} \\
        \midrule
        TTUR=1:5 + IDA & 4.25 & 4.20 \\
        TTUR=1:5 + $\mathcal L_\mathrm{CD}$ & 5.51 & 4.67 \\
        \bottomrule
    \end{tabular}
    }
\end{minipage}
\end{minipage}
\end{figure}

\textbf{Hyperparameters.} For $\kappa$ in the flow-map-initialized setting, the ablation in \cref{tab:ab-gamma} shows that timestep shifting improves distribution matching compared to the unshifted setting ($\kappa=1$). We then fix $\kappa=10$ and study $\lambda$ with $w_t$ in \cref{tab:ab-lambda}. Adaptive scaling gives better FID for $\lambda \ge 0.05$. Thus, we use $\lambda \times w_t = 0.1 / |\bar u_\theta(z; 1, 0) - v_\Phi(\hat x_t; t)|$ in the following experiments. We also evaluate $\lambda \times w_t=0.5 \times 1.0$ while clipping $\Delta_{u, \Phi}$ to $[-1, 1]$, following \cite{cheng2026twinflow}. This gives an FID of 3.98, so we keep the adaptive scaling. For the flow-matching-initialized setting, adding the cosine factor to $w_t$ improves FID from 4.82 to 4.70. We study TTUR ratios and the ablation of the GAN loss in \cref{tab:dmd2-sweep}, and report the best FID in \cref{tab:main-comparison}. In this sweep, FSF-DMD obtains a lower FID than the listed explicit-network variants at the evaluated TTUR ratios.

\textbf{Random masking} is a regularization factor for consistency distillation, and we do not add it as a DMD2 factor in the main comparison. For $\mathcal L_\mathrm{CD}$ jointly trained with DMD2, we note that the masking strategy also improves FID from 5.51 to 4.99 without GAN, and from 4.67 to 3.85 with GAN. Since this case requires 1.68 sec/step and 76GB VRAM, FSF-DMD provides an effective and efficient alternative without an additional network, a separate training loop, or GAN. We also note that for the flow-matching-initialized setting, random masking degraded the performance of the DMD2 + $\mathcal L_\mathrm{CD}$ baseline (FID 4.93). We therefore retain the better unmasked result in \cref{tab:ldit-teacher}.

\textbf{Stop-Gradient Placement.} We study the effect of the additional stop-gradients according to batch indices. Since the resulting pipelines are identical when $M=1$ with or without the stop-gradient, we focus on $M=2$. With the vanilla backward simulation for flow-map generators in \cref{eq:tilde-F}, the FID is 3.85 with a VRAM requirement of 68GB. After applying additional stop-gradients according to batch indices, each sample retains gradients through one of the $M$ sampling steps. This preserves the FID of 3.85 while reducing the VRAM requirement to 49GB, matching that of $M=1$. We observe consistent performance across both setups and thus report the latter for the $M=2$ settings.

\textbf{Analysis.} As shown in \cref{tab:fsfdmd-wo-cd}, we observe the behavior of FSF-DMD without consistency distillation $\mathcal L_\mathrm{CD}$. FSF-DMD begins to diverge after 1K training steps when $\mathcal L_\mathrm{CD}$ is omitted. One possible explanation is that the pseudo-velocity surrogate depends on flow-map properties, including point-mass collapse, injectivity, and invertibility. Without this anchor, the objective pushes $u_\theta(\hat x_t; t, 0)$ toward $v_\Phi(\hat x_t; t)$, and one-step generation may collapse toward the data mean due to the posterior-mean property~\citep{karras2022elucidating}. Thus, we jointly train FSF-DMD with consistency distillation.

\begin{table}[th]
\centering
\caption{FSF-DMD without consistency distillation $\mathcal L_\mathrm{CD}$ in the flow-map-initialized setting.}
\label{tab:fsfdmd-wo-cd}
\vspace{1em}
\begin{tabular}{l|cccc}
    \toprule
    Training steps & 1K & 2K & 4K & 6K \\
    \midrule
    FID & 4.04 & 6.61 & 13.84 & 24.03 \\
    \bottomrule
\end{tabular}
\end{table}

Since the perturbation process is altered, the teacher- and fake-velocity fields can be inconsistent. To quantify this effect, we measure the cosine similarity and MSE between our surrogate and the conventionally trained fake velocity in \cref{tab:vel-discrepancy}. The generator is trained with FSF-DMD in the flow-map-initialized setting, and the auxiliary fake velocity follows our DMD2 training protocol without affecting the training process of FSF-DMD. MSE (\%) denotes the MSE normalized by the squared norm of the explicitly trained fake-velocity estimate, expressed as a percentage.

\begin{table}[th]
\centering
\caption{Discrepancy between the surrogate and estimated fake velocity over training steps.}
\label{tab:vel-discrepancy}
\vspace{1em}
\begin{tabular}{ccccc}
    \toprule
    Training steps & Cosine sim. & MSE & MSE (\%) & FID \\
    \midrule
    0K - 0.2K & 0.93 & 0.38 & 14\% & 5.57 \\
    0.2K - 1K & 0.51 & 12 & 79\% & 4.49 \\
    1K - 2K & 0.41 & 18 & 90\% & 4.12 \\
    2K - 3K & 0.41 & 18 & 90\% & 3.89 \\
    3K - 4K & 0.42 & 18 & 88\% & 3.85 \\
    \bottomrule
\end{tabular}
\end{table}

Initially, the cosine similarity is higher and the MSE is lower than at later stages, and the surrogate is partly aligned with the usual fake velocity. After 200 steps, the similarity decreases, while FSF-DMD obtains FIDs comparable to those of DMD variants in \cref{tab:main-comparison}. Together with \cref{tab:ldit-teacher} and \cref{tab:from-scratch}, our surrogate remains effective in practice while maintaining computational efficiency.

As one possible intuitive explanation, the update induced by \cref{eq:fsf-dmd-ideal} could provide a meaningful teacher-directed correction to generated samples relative to their expected generative direction, $u_\theta(\hat x_t; t, 0)$. This could steer the generative direction toward the teacher velocity through a DMD-style correction, complementing flow-map training. Such a correction could remain useful without requiring the heuristic surrogate to be the exact fake velocity. We did not observe unusual optimization behavior attributable to the change of perturbations in our settings (\cref{fig:loss-curves}).

A promising direction is to study how these differences affect final generation quality, and we leave this for future work.

\begin{table}[th]
\centering
\caption{\textbf{Additional metrics for ImageNet-1K $\bf 256\!\times\!256$ comparisons.} We report FID50K, Inception Score (IS), and precision/recall at the best-FID checkpoint for each listed method.}
\label{tab:additional-metrics}
\vspace{1em}
\resizebox{0.95\textwidth}{!}{%
\begin{tabular}{lcrrrcc}
\toprule
Method & sec/step & Peak step & FID ($\downarrow$) & IS ($\uparrow$) & Precision ($\uparrow$) & Recall ($\uparrow$) \\
\midrule
\multicolumn{7}{l}{Flow-map-initialized setting} \\
\midrule
Consistency distillation & 0.23 & 36K & 6.06 & 188.90 & 0.65 & 0.47 \\
DMD2 & 1.46 & 15K & 4.18 & 237.12 & 0.60 & 0.50 \\
DMD2 + IDA-style update & 1.46 & 12K & 4.20 & 244.09 & 0.60 & 0.49 \\
DMD2 + $\mathcal L_\mathrm{CD}$ & 1.63 & 54K & 4.67 & 221.87 & 0.61 & 0.49 \\
\midrule
\textbf{FSF-DMD (Ours, $M=2$)} & 0.38 & 4K & 3.85 & 262.50 & 0.59 & 0.53 \\
\textbf{FSF-DMD (Ours, $M=1$)} & 0.31 & 3K & 3.96 & 207.83 & 0.62 & 0.51 \\
\midrule
\multicolumn{7}{l}{Flow-matching-initialized setting} \\
\midrule
Consistency distillation & 0.11 & 19K & 5.53 & 280.90 & 0.48 & 0.53 \\
DMD2 (TTUR=1:5) & 0.82 & 3K & 5.53 & 287.33 & 0.47 & 0.54 \\
DMD2 + $\mathcal L_\mathrm{CD}$ (TTUR=1:5) & 0.90 & 33K & 4.90 & 262.03 & 0.54 & 0.53 \\
\midrule
\textbf{FSF-DMD (Ours)} & 0.17 & 8K & 4.70 & 246.01 & 0.55 & 0.54 \\
\bottomrule
\end{tabular}
}
\end{table}

\begin{figure}[th]
\centering
\includegraphics[width=0.8\linewidth]{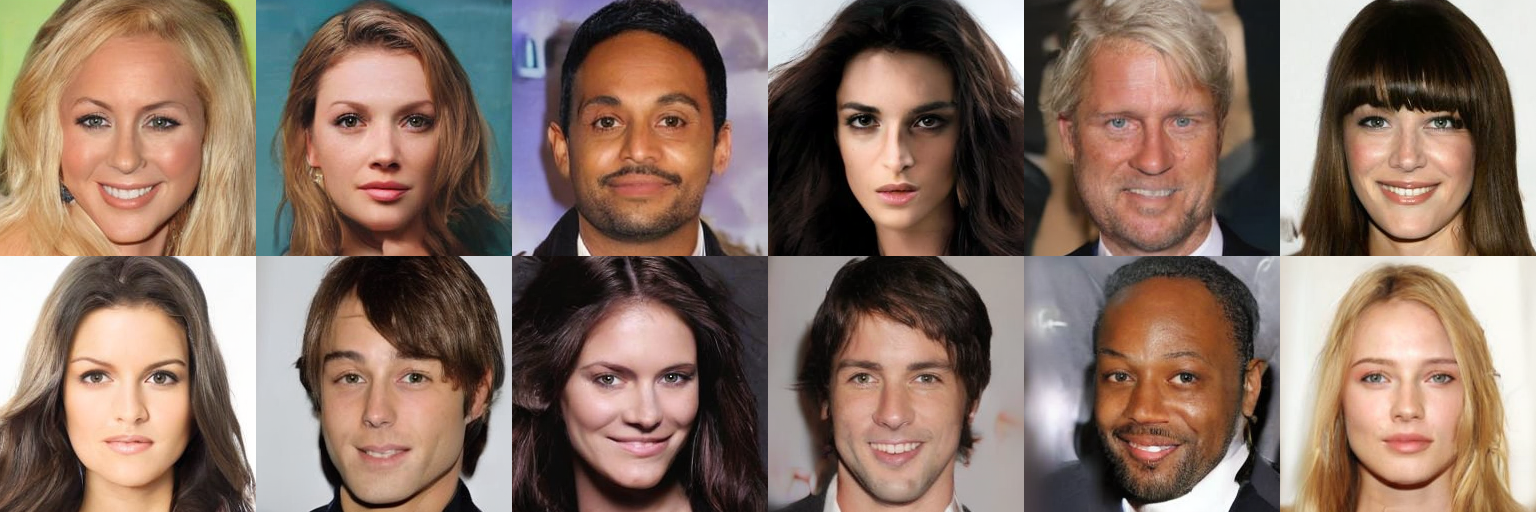}
\caption{Qualitative results of FSF-DMD trained on CelebA-HQ.}
\label{fig:celeba-hq}
\end{figure}

\section{Impact Statement}
\label[appendix]{app:impact-statement}

This work aims to advance machine learning by improving the efficiency of distribution matching for flow-map generative models. Its contributions are focused on analyzing the fake-side term in DMD and reformulating it with a generator-induced pseudo-velocity surrogate, without introducing an additional fake-score network. The ethical and societal implications are therefore largely aligned with those of existing generative-modeling research. These include potential concerns around dataset bias, misuse of generated content, and downstream applications. On the positive side, reducing the need for auxiliary fake-score networks may improve research efficiency and reduce computational cost. The paper does not directly address broader societal risks, which remain important considerations for future work.

\section{Licenses}
\label[appendix]{app:licenses}

This work uses the following existing codebases, models, and datasets:
\begin{itemize}
  \item LightningDiT and VA-VAE~\citep{yao2025vavae} (\url{https://github.com/hustvl/LightningDiT}): Used as the backbone implementation and latent preprocessor for the ImageNet experiments. Distributed under the MIT License.
  \item SD-VAE~\citep{Rombach_2022_CVPR} (\url{https://huggingface.co/stabilityai/sd-vae-ft-mse}): Used as the latent preprocessor for the CelebA-HQ experiments. Distributed under the MIT License.
  \item ImageNet-1K Dataset: Used for image-generation experiments. Licensed for non-commercial research and educational purposes.
  \item CelebA-HQ resized ($256\!\times\!256$) Dataset: Used for image-generation experiments. Licensed under the GPL-2.0 License.
  \item UCGM (\url{https://github.com/LINs-lab/UCGM}): Used as a training-loop baseline for the ImageNet experiments. Licensed under the Apache-2.0 License.
  \item DMD2~\citep{yin2024improved} (\url{https://github.com/tianweiy/DMD2}): Used as a reference implementation for the DMD2 baseline. Licensed under CC BY-NC-SA 4.0.
  \item SenseFlow~\citep{ge2026senseflow} (\url{https://github.com/XingtongGe/SenseFlow}): Used as a reference implementation for Implicit Distribution Alignment (IDA). Licensed under the Apache-2.0 License.
  \item rCM~\citep{zheng2026large} (\url{https://github.com/NVlabs/rcm}): Used as a reference implementation for the rCM baseline. Licensed under the Apache-2.0 License.
  \item TwinFlow~\citep{cheng2026twinflow} (\url{https://github.com/inclusionAI/TwinFlow}): Used as a reference implementation for the TwinFlow and RCGM baselines. Licensed under the Apache-2.0 License.
  \item DMF~\citep{lee2026decoupled} (\url{https://github.com/kyungmnlee/dmf}): Used as a reference implementation for Decoupled MeanFlow.
  \item StyleGAN3 (\url{https://github.com/NVlabs/stylegan3}): Used for computing precision and recall metrics. Licensed under the NVIDIA Source Code License for StyleGAN3.
\end{itemize}

\FloatBarrier

\end{document}